\documentclass{LMCS}

\def\eqalign#1{\null\,\vcenter{\openup\jot\mathsurround=0 pt
  \ialign{\strut\hfil$\displaystyle{##}$&$\displaystyle{{}##}$\hfil
      \crcr#1\crcr}}\,}

\usepackage{url}
\usepackage{enumerate}
\usepackage{latexsym}
\usepackage{amsfonts}
\usepackage{mathrsfs}
\usepackage{epsfig}
\usepackage{hyperref}

\newtheorem{theorem}{Theorem}[section]

\newtheorem{proposition}[theorem]{Proposition}
\newtheorem{lemma}[theorem]{Lemma}

\newtheorem{example}[theorem]{Example}
\newtheorem{definition}[theorem]{Definition}
\newtheorem{remark}[theorem]{Remark}

\long\def\COMMENT#1\ENDCOMMENT{\message{(Commented text...)}\par}

\def\st{\smallskip \noindent}

\def\and{ \ \wedge}

\def\iff{\;\Leftrightarrow\;}

\def\beq{\begin{equation}}
\def\eeq#1{\label{#1}\end{equation}}

\def\cala{{\mathcal A}}

\def\calS0{{\mathcal S}_0}

\def\lan{\langle}
\def\ran{\rangle}

\def\doi{2 (4:2) 2006}
\lmcsheading%
{\doi}
{1--34}
{}
{}
{Jan.~13, 2006}
{Oct.~\phantom{0}1, 2006}
{}

\begin{document}

\title[State-Based Regression in Incomplete Domains]
{A State-Based Regression Formulation for
Domains with Sensing Actions and Incomplete Information}

\author[L.-C.~Tuan]{Le-Chi Tuan\rsuper{a}}
\address{{\lsuper{a}}
GCAS Incorporated, 1531 Grand Avenue, San Marcos, CA 92078, USA}
\email{lctuan@gcas.net}

\author[C.~Baral]{Chitta Baral\rsuper{b}}
\address{{\lsuper{b}}
Computer Science and Engineering,
Arizona State University, Tempe, AZ 85287, USA}
\email{chitta@asu.edu}

\author[T.~C.~Son]{Tran Cao Son\rsuper{c}}
\address{{\lsuper{c}}
Computer Science Department,
New Mexico State University,
Las Cruces, NM 88003, USA}
\email{tson@cs.nmsu.edu}



\begin{abstract}
We present a \emph{state-based regression function} for planning
domains where an agent does not have complete information and may
have sensing actions. We consider  binary domains and employ a
three-valued characterization of domains with sensing actions to
define the regression function. We prove the soundness and
completeness of our regression formulation with respect to the
definition of progression. More specifically, we show that (i) a
plan obtained through regression for a planning problem is indeed
a progression solution of that planning problem, and that (ii) for
each plan found through progression, using regression one obtains
that plan or an equivalent one.
\end{abstract}

\maketitle

\section{Introduction and Motivation}

An important aspect in reasoning about actions and characterizing
the semantics of action description languages is to define a
transition function that encodes the transition between states due
to actions. This transition function is often viewed as a {\em
progression} function in that it denotes the progression of the
world by the execution of actions. The `opposite' or `inverse' of
progression is referred to as {\em regression}.

  Even for a simple case where we have only non-sensing actions
and the progression transition function is deterministic, there are
various formulations of regression. For example, let us consider the
following. Let $\Phi$ be the progression transition function from
actions and states to states. I.e., intuitively, $\Phi(a,s) = s'$
means that if the action $a$ is executed in the state $s$ then the
resulting state will be $s'$. One way to define a regression
function $\Psi_1$ is to define it with respect to states. In that
case $s \in \Psi_1(a,s')$ will mean that the state $s'$ is reached
if $a$ is executed in $s$. Another way to define regression is with
respect to formulas. In that case $\Psi_2(a,f) = g$, where $f$ and
$g$ are formulas, means that if $a$ is executed in a state
satisfying $g$ then a state satisfying $f$ will be reached.

  For planning using heuristic search, often a different
formulation of regression is given. Typically, a planning problem is
specified by a set of actions, an initial state, and a goal state,
which is a conjunction of literals. As such, regression is often
defined with respect to a set of literals and an action. In that
case the conjunction of literals (the goal) denotes a set of states,
one of which needs to be reached. This regression is slightly
different from $\Psi_2$ as the intention is to regress to another
set of literals (not an arbitrary formula), denoting a sub-goal.

  With respect to the planning language STRIPS \cite{fik71}, where each
action $a$ has an add list $Add(a)$, a delete list $Del(a)$, and a
precondition list $Prec(a)$, the progression function is defined
as $Progress(s,a) = s + Add(a) - Del(a)$; and the regression
function is defined as $Regress(conj,a) = conj + Prec(a) -
Add(a)$, where $conj$ is a set of atoms. Intuitively,
$Regress(conj,a)$ represents a minimal requirement on states
from which the execution of $a$ leads to states satisfying
$conj$. The relation between
these two, formally proven in \cite{ped86}, shows the correctness
of regression based planners; which, through use of
heuristics (e.g. \cite{bonet:geffner:hsp-01,nguyenetal02}), have done
well in planning competitions. However, the focus of these papers
has been the regression function in domains where agents have
{\em complete} knowledge about the world. The following example
shows that this property does not always holds.

\begin{example}
\label{ex-01}
Consider the following do-or-die story\footnote{
  This story was brought to us by a participant of
  a Texas Action Group (TAG) meeting at Lubbock in 2002 during a discussion
  on the need of sensing actions in reasoning about actions
  and changes and planning.
}:

\begin{itemize}
\item[] A {\em wannabe prince} faces the last task in
his endeavor. He stands in front of two rooms. In one room is a
tiger and in the other is the princess, whom he wants to marry. Opening
the room with the tiger will result in him being eaten. Otherwise,
he will be able to rescue the princess and will get to marry her. He
does not know exactly in which room the princess is. However, he can
use a specialized\footnote{Because the rooms are too close to each
other, the natural smelling ability of a human is not quite
accurate.} smell sensor that can precisely tell him where the tiger is.
\end{itemize}

The story can be formalized as follows. Let us denote the rooms by
$1$ and $2$. $in(t,R)$ (resp. $in(p,R)$) denotes that the tiger
(resp. the princess) is in room $R$. Initially, the agent (i.e., the
want-to-be prince) is $alive$; he does not know what is behind the
door of each room (i.e., the truth value of $in(t,R)$ and $in(p,R)$
is unknown to him) but he knows that the tiger and the princess are
in different rooms (i.e., if $in(t,1)$ is true then $in(p,2)$ is
true, etc.); he can execute $open(1)$ and $open(2)$. Executing the
action $open(R)$, when $in(t,R)$ is true, causes him to be death
($\neg alive$); otherwise, the princess gets rescued. The agent can
{\em determine} (by smelling) the truth value of $in(t,1)$ and
$in(t,2)$. If he dies, he can not execute any action.

It is easy to see that the only possible way for the agent to achieve
his goal is to begin by determining where the tiger is (by executing
the action {\em smell}); after that, depending on where the tiger is,
he can open the other room to rescue the princess.  Observe that this
plan involves the action $smell$ whose execution does not change the
world but changes the knowledge of the agent.  Furthermore, the second
action of the plan depends on the knowledge of the agent after the
execution of the first action. We say that the agent needs a {\em
conditional plan} with sensing actions to achieve his goal.
\end{example}

Reasoning about the effects of actions and changes in the presence
of sensing actions and incomplete information has been the topic of
intensive research (e.g.,
\cite{lev96,lobo97,moo85b,rei01b,sonbaral00} and the discussion in
these papers). In general, the progression function for action
theories with sensing actions and incomplete information is defined
as a mapping from pairs of actions and belief states to belief
states, where each belief state is a set of possible states.
Intuitively, a belief state represents the set of possible states an
agent thinks he might be in given his knowledge about the world. For
example, the initial belief state of the want-to-be-prince in
Example \ref{ex-01} consists of every possible state of the world;
and, after the execution of the action {\em smell}, his belief state
consists of a single state in which he is alive and knows the
location of the tiger and the princess.

It has been also recognized that the planning problem in domains
with sensing actions and incomplete information has a higher
complexity than the planning problem in domains with complete
information \cite{baral:ijcai99:aij}. Furthermore, plans for
achieving a goal in these domains will sometime require sensing
actions and conditionals \cite{lev96,sonbaral00}. It should be noted that
there is an alternative approach to planning in the presence of
incomplete information, called conformant planning, where no sensing
action is used and a plan is a sequence of actions leading to the
goal from every possible initial situation. Example \ref{ex-01}
indicates that this is inadequate for many planning problems. In the
past, several planners capable of generating conditional plans have
been developed (e.g., \cite{bryceKS04,lob98,TuSB06}) in which some
form of the progression function has been used.

In this regards, two natural questions arise:

\begin{enumerate}[$\bullet$]

\item How to define a regression function in the presence of
incomplete information and sensing actions?

\item What should be the result of the regression of a
state or a formula over a conditional plan? and, how can it be computed?

\end{enumerate}

In the literature, we can find several proposals addressing the
first question \cite{ScherlL03,rei01b,deg99,sonbaral00},  among them
only the proposal in \cite{sonbaral00} partly discusses the second
one. Moreover, all previous regression functions with respect to
domains with incomplete information and sensing actions are about
regression of formulas.

{\em In this paper we are concerned with domains where the agent
does not have complete information about the world, and may have
sensing actions.} For such domains, we define a regression function
with respect to states. We then formally relate our definition of
regression with the earlier notion of progression and show that
planning using our regression function will not only give us correct
plans but also will not miss plans. In summary the main
contributions of our paper are:

\begin{enumerate}[$\bullet$]

\item  A state-based regression function
for STRIPS domains with sensing actions and incomplete
initial state;

\item An extended regression function that allows for
the regression from a (goal) state over a conditional plan;
and

\item A formal result showing the soundness
and completeness of our regression function with respect to the
progression function.

\end{enumerate}

The rest of this paper is organized as follows. First, we review
the necessary background information for understanding the
technical details of the paper. We then present the regression
formulation (Section \ref{Reg})
and prove its soundness and completeness with respect
to the progression function (Section \ref{result}).
We relate
our work to other work in regression in Section \ref{related} and conclude in
Section \ref{conc}.

\section{Background}

In this section, we present our action and plan representation
and its semantics.

\subsection{Action and Plan Representation} \label{pre}

We employ a STRIPS-like action representation~\cite{fik71} and
represent a planning problem by a tuple $P= \langle A,O,I,G
\rangle$ where $A$ is a finite set of fluents, $O$ is a finite set
of actions, and $I$ and $G$ are sets of fluent literals
where a fluent literal is either a fluent $f \in A$ (a.k.a.
{\em positive fluent literal}) or its negation $\neg f$ (a.k.a.
{\em negative fluent literal}).
Intuitively, $I$ encodes what is known about the initial state and
$G$ encodes what is desired of a goal state.

  An action $a \in O $ is either a {\em non-sensing action} or a
{\em sensing action} and is defined as follows:
\begin{enumerate}[$\bullet$]
     \item
     A non-sensing action $a$ is specified by an expression of the form
\[
\textnormal{action }
a
\hspace*{.2in}
\textnormal{:Pre } Pre_a
\hspace*{.2in}
\textnormal{:Add } Add_a
\hspace*{.2in}
\textnormal{:Del } Del_a
\]
     where $Pre_a$ is a set of fluent literals representing the
     precondition for $a$'s execution, $Add_a$  and
     $Del_a$ are two disjoint sets of fluents representing
     the positive and negative effects of $a$, respectively; and

     \item A sensing action $a$ is specified by an expression of the form
\[
\textnormal{action }
a
\hspace*{.2in}
\textnormal{:Pre }
Pre_a
\hspace*{.2in}
\textnormal{:Sense }Sens_a
\]
     where $Pre_a$ is a set of fluent literals and $Sens_a$ is a subset of
     the fluents that do not appear in $Pre_a$, .i.e.,
     $Pre_a \cap (\{\neg f \mid f \in Sens_a\} \cup Sens_a) = \emptyset$.
     As with non-sensing actions, a sensing action might only be executed
     under certain condition, which is represented by $Pre_a$.
     Intuitively, $Pre_a$ is the condition under which $a$
     can be executed, and hence, needs to be known to be true
    before the execution
     of $a$. On the other hand, $Sens_a$ is the set of fluents that
     are unknown at the time of execution. For this reason, we require
     that none of the fluents in $Sens_a$ appear in $Pre_a$.
     As an example of a sensing action with precondition,
     consider the action of looking into
     the refrigerator to determine whether there is some beer or not.
     This action requires that the refrigerator door is open and
     can be represented by the action $look$ with the condition
     $Pre_{look} = \{door\_open\}$ and $Sens_{look} = \{beer\_in\_fridge\}$.

\end{enumerate}
The next example shows a simple domain  in our representation.

\begin{example}
Figure (\ref{example-fig01}) shows the actions of
the ``Getting to Evanston'' domain from~\cite{pry96}
in our representation.

\begin{figure}[h]
\centerline{
\begin{tabular}{|l|l|l|l|} \hline \hline
\multicolumn{1}{|c|} {{\bf Action Name}} &
\multicolumn{1}{|c|} {{\bf :Pre}} & \multicolumn{1}{|c|} {{\bf
:Add}} & \multicolumn{1}{|c|} {{\bf :Del}} \\ \hline \hline
goto-western-at-belmont & \{at-start\} & \{on-western,  & \{at-start\}\\
                       &  &  \ \ \ \ on-belmont\} & \\
take-belmont & \{on-belmont,  & \{on-ashland\} & \{on-western\}\\
             &  \ \ \ \ \ \ traffic-bad\} &  & \\
take-ashland & \{on-ashland\} & \{at-evanston\} & \\
take-western & \{$\neg$traffic-bad,  & \{at-evanston\} & \\
              &  \ \ \ \ \ on-western\} &  & \\
\hline \hline \multicolumn{1}{|c|} {{\bf Action Name}} &
\multicolumn{1}{|c|} {{\bf :Pre}} & \multicolumn{2}{|c|} {{\bf
:Sense}} \\ \hline \hline
\multicolumn{1}{|c|} {check-traffic} & \multicolumn{1}{|c|} {$\emptyset$} & \multicolumn{2}{|c|} {\{traffic-bad\}} \\
\multicolumn{1}{|c|} {check-on-western} & \multicolumn{1}{|c|}
{$\emptyset$} & \multicolumn{2}{|c|} {\{on-belmont\}} \\ \hline
\end{tabular}
}
\normalsize
\caption{Actions of the ``Getting to Evanston'' domain.}
     \label{example-fig01}
\end{figure}
The first four rows of the tables describe different non-sensing
actions (driving actions) with their corresponding preconditions and
add- and delete-effects. Each action can be executed when the agent
is at certain locations (the second column) and changes the location
of the agent after its completion. For instance, {
goto-western-at-belmont} can be executed if the agent is {
at-start}; its effect is that the agent will be { on-western} and {
on-belmont} (third column) and no longer { at-start} (fourth
column).

The last two rows represent two sensing actions, neither requires a
precondition; one allows the agent to check for traffic condition ({
check-traffic}) and the other one ({ check-on-western}) allows for
the agent to check whether it is ({ on-belmont}) or not.
\end{example}

The notion of a plan in the presence of incomplete information and
sensing actions has been extensively discussed in the
literature~\cite{lev96,lobo97,ScherlL03,sonbaral00}. In this paper,
we consider {\em conditional plans} that are formally defined as
follows.

\begin{definition} [Conditional Plan] \label{defC}\hfill

\begin{enumerate}[$\bullet$]
     \item An empty sequence of actions, denoted by $[\ ]$, is a conditional plan.
     \item If $a$ is a non-sensing action, then $a$ is a conditional plan.
     \item If $a$ is a sensing action and
     $\varphi_1,\ldots,\varphi_n$ are mutually exclusive
     conjunctions of fluent literals
     and $c_1, \ldots, c_n$ are conditional plans, then
\[
a; case(\varphi_1 \rightarrow c_1, \ldots, \varphi_n \rightarrow c_n)
\]
  is a conditional plan \footnote{We often refer to this type of conditional plans as \emph{case
  plans
         }.}.
     \item If
          $a$ is a non-sensing action and
$c$ is a conditional plan, then $a;c$ is a
         conditional plan.
     \item Nothing else is a conditional plan.
\end{enumerate}
\end{definition}
  Intuitively, to execute a plan $a; case(\varphi_1 \rightarrow
c_1, \ldots, \varphi_n \rightarrow c_n)$, first $a$ is executed,
$\varphi_i$'s are then evaluated. If one of $\varphi_i$ is true
then $c_i$ is executed. If none of $\varphi_i$ is true then the
plan fails. To execute a plan $a; c$, first $a$ is executed
then $c$ is executed.

\begin{example} [Getting to Evanston] \label{ex0}
The following is a conditional plan:\\
  \hspace*{0.25in} $\mathit{check\mbox{-}traffic};\\
\hspace*{0.3in} case(\ \\
\hspace*{0.5in} \mathit{traffic\mbox{-}bad} \rightarrow \\
\hspace*{0.7in}  goto\mbox{-}western\mbox{-}at\mbox{-}belmont;\\
\hspace*{0.7in}  take\mbox{-}belmont;\\
\hspace*{0.7in}  take\mbox{-}ashland\\
\hspace*{0.5in} \neg \mathit{traffic\mbox{-}bad} \rightarrow \\
\hspace*{0.7in}  goto\mbox{-}western\mbox{-}at\mbox{-}belmont; \\
\hspace*{0.7in} take\mbox{-}western\\
\hspace*{0.3in} ) $ 
\end{example}

\subsection{The Progression Function}
\label{sub1}

In the presence of incomplete information, the knowledge of an agent
can be approximately captured by three disjoint sets of fluents: the
set of fluents known to be true, false, and unknown to him,
respectively. Thus, we represent the knowledge of an agent by a pair
$\langle T, F \rangle$, called an {\em approximate state} (or
a-state), where $T {\subseteq} A$ and $F {\subseteq} A$ are two
disjoint sets of fluents. Intuitively, $\langle T, F \rangle$
represents the knowledge of an agent
    who knows that
    fluents in $T$ (resp. $F$) are true (resp. false) and
     does not have any knowledge about fluents in $A \setminus (T \cup F)$.
It can also be considered as the
     intersection of all belief states satisfying
     $T \cup \{\neg f \mid f \in F\}$.

Given a fluent $f$, we
     say that $f$ is true (resp. false) in $\sigma$ if $f
     \in T$ (resp. $f \in F$). $f$ (resp. $\neg f$) holds in $\sigma$
     if $f$ is true (resp. false) in $\sigma$. $f$ is known (resp.
     unknown) in $\sigma$ if $f \in (T \cup F)$ (resp. $f \not \in (T
     \cup F)$). A set $L$ of fluent literals holds in an a-state
     $\sigma = \langle T,F \rangle$ if every member of $L$ holds in
     $\sigma$. A set $X$ of fluents is known in $\sigma$ if every
     fluent in $X$ is known in $\sigma$. An action $a$ is {\em
     executable} in $\sigma$ if $Pre_a$ holds in $\sigma$.
Furthermore, for two a-states  $\sigma_1 {=} \langle T_1,F_1 \rangle$ and $\sigma_2 {=}
     \langle T_2,F_2 \rangle$:

     \begin{enumerate}[(1)]
         \item We call $\sigma_1 {\cap} \sigma_2 {=} \langle T_1 {\cap} T_2, F_1 {\cap} F_2 \rangle$
         the intersection of $\sigma_1$ and $\sigma_2$.
         \item We say $\sigma_1$ extends $\sigma_2$, denoted by $\sigma_2 {\preceq}
         \sigma_1$ if $T_2 {\subseteq} T_1$ and $F_2 {\subseteq} F_1$.
         $\sigma_1 {\setminus} \sigma_2$ denotes the set $(T_1 {\setminus}
         T_2) {\cup} (F_1 {\setminus} F_2)$.
         \item For a set of fluents $X$, we
         write $X {\setminus} \langle T,F \rangle$ to denote $X {\setminus}
         (T {\cup} F)$. To simplify the presentation, for a set of literals
         $L$, by $L^+$ and $L^-$ we denote the set of fluents $\{f \mid
         f{\in}L,\; f $ is a fluent $\}$ and  $\{f \mid \neg f{\in}L,\; f $
         is a fluent $\}$.
     \end{enumerate}

The transition function (for progression) is defined next.

\begin{definition} [Transition Function] \label{def-tran}
For an a-state $\sigma=\langle T,F \rangle$ and an action $a$,
$\Phi(a,\sigma)$ is defined as follows:
\begin{enumerate}[$\bullet$]
   \item  if $a$ is not executable in $\sigma$ then $\Phi(a,\sigma) =
   \{\bot\}$;
   \item  if $a$ is executable in $\sigma$ and $a$
    is a non-sensing action then
     $$\Phi(a,\sigma) =
\{\langle (T \setminus Del_a) \cup Add_a, (F \setminus Add_a) \cup
Del_a \rangle\};$$
   \item  if $a$ is executable in $\sigma$ and $a$ is a sensing action then
     $$\Phi(a,\sigma) = \{\sigma' | \sigma \preceq \sigma' \textnormal{ and }
     Sens_a \setminus \sigma = \sigma' \setminus \sigma\}.$$
\end{enumerate}
\end{definition}

Here, $\bot$ denotes the ``error state.''
$\Phi(a,\sigma) = \{\bot\}$ indicates that the action $a$ cannot be
executed in the a-state $\sigma$.
The next example illustrates the above definition.

\begin{example}
[Getting to Evanston] \label{ex2} Consider the a-state
\[\sigma=\langle \{at\hbox{-}start\},\{on\hbox{-}western,
on\hbox{-}belmont,on\hbox{-}ashland, at\hbox{-}evanston \}
\rangle.\]
We have that $\mathit{check\mbox{-}traffic}$ is executable in $\sigma$
and
\[\Phi(\mathit{check\mbox{-}traffic}, \sigma)= \{\sigma_1, \sigma_2\}\] where:\\
$\sigma_1 = \langle \{at\hbox{-}start, \mathit{traffic\hbox{-}bad}\},
\{on\hbox{-}western, on\hbox{-}belmont, on\hbox{-}ashland,
at\hbox{-}evanston \} \rangle$,\\ $\sigma_2 = \langle
\{at\hbox{-}start\},\{\mathit{traffic\hbox{-}bad}, on\hbox{-}western,
on\hbox{-}belmont, on\hbox{-}ashland, at\hbox{-}evanston\}
\rangle$.

  Similarly, \[\Phi(goto\mbox{-}western\mbox{-}at\mbox{-}belmont,\sigma) =
\{\sigma_3\}\] where: $\sigma_3 = \langle
\{$on\mbox{-}western,on\mbox{-}belmont$\}, \{$at\mbox{-}start, on\mbox{-}ashland,
at\mbox{-}evanston $\} \rangle$. 
\end{example}
  The function $\Phi$ can be extended to define the function
$\Phi^*$ that maps each pair of a conditional plan $p$ and
a-states $\sigma$ into a set of a-states, denoted by
$\Phi^*(p,\sigma)$. $\Phi^*$ is defined similarly to the extended function
$\hat\Phi$ in~\cite{sonbaral00}.

\begin{definition} [Extended Transition Function] \label{def-extran}
For an a-state $\sigma$,
\begin{enumerate}[$\bullet$]
     \item  if $c = [\ ]$, then $\Phi^*([\ ], \sigma) = \{\sigma\}$;

     \item  if $c = a$ and
            $a$ is a non-sensing action,
        then $\Phi^*(c, \sigma) = \Phi(a,\sigma)$;
     \item if $c=a;case(\varphi_1 \rightarrow p_1, \ldots, \varphi_n \rightarrow p_n)$ is a case plan,
     then
         \[\Phi^*(c, \sigma) = \bigcup_{\sigma' \in \Phi(a,\sigma)} E(case(\varphi_1 \rightarrow p_1, \ldots, \varphi_n \rightarrow p_n), \sigma')\] where
             \[ E(case(\varphi_1 \rightarrow p_1, \ldots, \varphi_n \rightarrow p_n), \gamma) =
             \left\{%
             \begin{array}{ll}
             \Phi^*(p_j,\gamma), & \hbox{if $\varphi_j$ holds in $\gamma$ ($1 \leq j \leq n$);} \\
             \{\bot\}, & \hbox{if none of $\varphi_1, \ldots, \varphi_n$ holds in $\gamma$.} \\
             \end{array}%
             \right.    \]

     \item if $c$ is a conditional plan and $a$ is a non-sensing action, then
           $$\Phi^*(a;c, \sigma) = \bigcup_{\sigma' \in \Phi(a,\sigma)} \Phi^*(c, \sigma').$$
\end{enumerate}
Furthermore, $\Phi^*(c, \bot) = \{\bot\}$ for any conditional plan $c$.
\end{definition}
Intuitively, $\Phi^*(c,\sigma)$ is the set of a-states
resulting from the execution of the plan $c$ in $\sigma$.

  Given a planning problem  $P=\langle A,O,I,G \rangle$, the
a-state representing $I$ is defined by $\sigma_I = \langle I^+,
I^- \rangle $. $\Sigma_G = \{\sigma \mid \sigma_G \preceq
\sigma\}$, where $\sigma_G = \langle G^+, G^- \rangle$, is the set
of a-states satisfying the goal $G$. We define a progression solution as
follows.

\begin{definition}[Progression Solution]
\label{prog-sol}
A {\em progression solution}
to the planning problem $P$ is a conditional
plan $c$ such that
$\Phi^*(c,\sigma_I) \subseteq \Sigma_G$.
\end{definition}

Note that, since $\bot$
is not a member of $\Sigma_G$, we have that $\bot \not\in
\Phi^*(c,\sigma_I)$ if $c$ is a progression solution to $P$.
In other words, the execution of $c$ will not fail if $c$
is a progression solution of $P$.

{\example [Getting to Evanston - cont'd] \label{ex1}
Let $P = \langle A,O,I,G \rangle$
where $A$ and $O$ are given in Figure  (\ref{example-fig01}) and

  \hspace*{0.25in} $I = \{at$-$start, \neg on$-$western, \neg
on$-$belmont, \neg on$-$ashland, \neg at$-$evanston\}$;

  \hspace*{0.25in} $G = \{at$-$evanston\}$,

\noindent respectively. We can easily check that the conditional plan
in Example \ref{ex0} is a progression solution of $P$. 
}

\subsection{Some Properties of the Progression Function $\Phi$}

There have been several proposals on defining a progression function
for domains with sensing actions and incomplete information
\cite{lev96,lobo97,moo85b,rei01b,sonbaral00}. We will show next that
for domains considered in this paper, the function $\Phi$
(Definition \ref{def-tran}) is equivalent to the transition function
defined in \cite{sonbaral00}. By virtue of the equivalent results
between various formalisms, in \cite{sonbaral00}, we can conclude
that $\Phi$ is equivalent to the progression functions defined in
several other formalisms as well. First, let us review the
definition of the function in \cite{sonbaral00}, which will be
denoted by $\Phi_c$. We need the following notations. For an action
theory given by a set of fluents $A$, a set of operators $O$, and an
initial state $I$, a state $s$ is a set of fluents. A {\em combined
state} (or {\em c-state}) of an agent is a pair $\langle s,\Sigma
\rangle$ where $s$ is a state and $\Sigma$ is a set of states.
Intuitively, the state $s$ in a c-state $\langle s,\Sigma \rangle$
represents the real state of the world whereas $\Sigma$ is the set
of possible states which an agent believes it might be in. A c-state
$\omega = \lan s,\Sigma \ran $ is {\em grounded} if $s \in \Sigma$.
A fluent $f$ is true (resp. false) in $s$ iff $f \in s$ (resp. $f
\not\in s$). $f$ is known to be true (resp. false)
in a c-state $\langle s,\Sigma
\rangle$ iff $f$ is true (resp. false) in every state $s' \in
\Sigma$; and $f$ is {\em known} in $\langle s,\Sigma \rangle$, if
$f$ is known to be true or known to be false in $\langle s,\Sigma
\rangle$.

For an action $a$ and a state $s$, $a$ is executable in $s$
if $Pre_a^+ \subseteq s$ and $Pre_a^- \cap s = \emptyset$. The state resulting from executing $a$
in $s$, denoted by $Res(a,s)$, is defined by
$Res(a,s) = (s \setminus Del_a) \cup Add_a$.
The function $\Phi_c$ is a mapping from pairs of
actions and c-states into c-states and is defined as
follows.  For a c-state $\omega = \langle s,\Sigma \rangle$
and action $a$,

\begin{enumerate}[(1)]

\item if $a$ is not
executable in $s$ then $\Phi_c(a, \omega)$ is undefined,
denoted by $\Phi_c(a, \omega) = \bot$;

\item if $a$ is executable in $s$ and $a$ is a non-sensing action,
then
$$\Phi_c(a,
\omega) = \langle Res(a,s), \{s'  \ | \   s' = Res(a, s''), \
    \exists s'' \in \Sigma \mbox{ s.t. } a \mbox{ is executable in }
  s''\} \rangle;
$$
and

\item if $a$ is executable in $s$ and $a$ is a
sensing action
then
$$
\Phi_c(a, \omega) = \langle s, \{s'  \ | \   s' \in \Sigma \mbox{ s.t. }
    Sens_a \setminus s = Sens_a \setminus s', \mbox{ and } a \mbox{ is executable in }
  s'\} \rangle.
$$
\end{enumerate}
The set of initial states of a planning problem $P = \langle A,O,I,G
\rangle$ is $\Sigma_0 = \{s \mid I^+ \subseteq s, \textnormal{ and }
I^- \cap s = \emptyset\}$; and the set of initial c-states of $P$,
denoted by $\Omega_I$, is given by $\Omega_I = \{\langle s_0,
\Sigma_0\rangle \mid s_0 \in \Sigma_0\}$. The function $\Phi_c$ can
be extended to define an extended progression function $\Phi^*_c$
over conditional plans and c-states, similar to the extended
function $\Phi^*$ in Definition \ref{def-extran}. The notion of a
progression solution can then be defined accordingly. The following
theorem states the equivalence between $\Phi$ and $\Phi_c$ for
domains representable by the action representation language given in the
previous subsection.

\begin{proposition} \label{prop1}
For a planning problem $\langle A, O, I, G\rangle$, a conditional
plan $c$ is a progression solution with respect to $\Phi$ iff it is
a progression solution with respect to $\Phi_c$.
\end{proposition}

\proof
 For an
a-state $\sigma = \langle T,F \rangle$, let $\Sigma_\sigma = \{s
\mid T \subseteq s \subseteq A \textnormal{ and } F \cap s =
\emptyset\}$, $\Delta_\sigma = \{\langle T',F' \rangle \mid T
\subseteq T' \subseteq A, \: F \subseteq F' \subseteq A,
\textnormal{ and } T' \cap F' = \emptyset\}$, and $\Omega_\sigma =
\{\langle s, \Sigma_\sigma)\rangle \mid s \in \Sigma_\sigma\}$.
Furthermore, for an action $a$ and a set of c-states $\Omega$,
let $\widehat{\Phi_c}(a, \Omega) =
                \{\Phi_c(a,\omega) \mid \omega \in \Omega\}$. From
the definition of $\Phi$ and $\Phi_c$, we can easily verify that the
following properties hold:
\begin{enumerate}[(1)]
\item An action $a$ is executable in $\sigma$ iff $a$ is executable in
every c-state belonging to $\Omega_\sigma$.
\item If a non-sensing action $a$ is executable in $\sigma$ and
$\Phi(a,\sigma) = \{\langle T',F'\rangle\}$ then
\[
T' = \bigcap_{u \in \Sigma, \langle s,\Sigma\rangle \in
                \widehat{\Phi_c}(a,\Omega_\sigma)} u
\]
and
\[
F' = \bigcap_{u \in \Sigma, \langle s,\Sigma\rangle \in
                \widehat{\Phi_c}(a,\Omega_\sigma)} (A \setminus u)
\]
\item If a sensing action $a$ is executable in $\sigma$ then
\[
\widehat{\Phi_c}(a,\Omega_\sigma) = \bigcup_{\sigma' \in
\Delta_\sigma, \: \sigma''
        \in \Phi(a, \sigma')} \Sigma_{\sigma''}
\]
\end{enumerate}
The conclusion of the proposition can be verified using induction on
the structure of a plan and the fact that for a planning problem
$P$, $\Omega_I = \Omega_{\sigma_I}$.\qed

\begin{remark}
{\rm For the discussion on the complexity of planning using the
progression function $\Phi$, it will be useful to note that $\Phi$
is also equivalent to the 0-approximation $\Phi_0$ defined in
\cite{sonbaral00}. Indeed, this can be easily verified from the
definitions of $\Phi_0$ and $\Phi$. In the notation of this paper,
$\Phi_0$ is also a mapping from pairs of actions and a-states into
a-states and for an a-state $\sigma=\langle T,F \rangle$ and an
action $a$, $\Phi_0(a,\sigma)$ is defined as follows:
\begin{enumerate}[$\bullet$]
   \item  if $a$ is not executable in $\sigma$ then $\Phi_0(a,\sigma) =
   \{\bot\}$;
   \item  if $a$ is executable in $\sigma$ and $a$
    is a non-sensing action then
     $$\Phi_0(a,\sigma) =
\{\langle (T \setminus Del_a) \cup Add_a, (F \setminus Add_a) \cup
Del_a \rangle\};$$
   \item  if $a$ is executable in $\sigma$ and $a$ is a sensing action then
     $$\Phi_0(a,\sigma) = \{\sigma' | \sigma \preceq \sigma' \textnormal{ and }
     Sens_a \setminus \sigma = \sigma' \setminus \sigma\}.$$
\end{enumerate}
This implies that $\Phi$ is identical to $\Phi_0$.
}
\end{remark}

\section{A State-Based Regression Formulation} \label{Reg}

In this section, we present our formalization of a regression
function, denoted by ${\mathcal R}$, and prove that it is both sound
and complete with respect to the progression function $\Phi$.
${\mathcal R}$ is a state-based regression function that maps each pair of
an action and a set of a-states into an a-state. 

Observe that our progression formulation states that a plan $p$
achieves the goal $G$ from an a-state $\sigma$ if $G$ holds in {\em
all} a-states belonging to $\Phi^*(p,\sigma)$, i.e., $G$ holds in
$\cap_{\sigma' \in \Phi^*(p,\sigma)} \sigma'$. In addition, similar
to \cite{bonet:geffner:hsp-01}, we will also define regression with
respect to the goal. These suggest us to introduce the notion of a
{\em partial knowledge state} (or p-state) as a pair $[T,F]$ where
$T \subseteq A$ and $F \subseteq A$ are two disjoint sets of
fluents. Intuitively, a p-state $\delta = [T,F]$ represents a
collection of a-states which extend the a-state $\langle T, F
\rangle$. We denote this set by $ext(\delta)$ and call it {\it the
extension set} of $\delta$. Formally, $ext(\delta) = \{\langle T',
F' \rangle \mid T \subseteq T' \subseteq A, F \subseteq F' \subseteq
A, T' \cap F' = \emptyset \}$. Any a-state $\sigma' \in ext(\delta)$
is called an extension of $\delta$. Given a p-state $\delta {=}
[T,F]$, we say a partial state $\delta' = [T',F']$ is a partial
extension of $\delta$ if $T \subseteq T', F \subseteq F'$. For a set
of p-states $\Delta=\{\delta_1,\ldots,\delta_n\}$, $\Delta'=
\{\delta'_1,\ldots,\delta'_n\}$ is said to be an extension of
$\Delta$, written as $\Delta \sqsubseteq \Delta'$ if $\delta'_i$ is
a partial extension of $\delta_i$ for every $i=1,\ldots,n$. For a
fluent $f$, we say that $f$ is true (resp. false, known, unknown) in
$\delta$ if $f \in T$ ($f \in F$, $f \in T \cup F$, $f \not\in T
\cup F$). A set of fluents $S$ is said to be true (resp. false,
known, unknown) in $\delta$ if every fluent $f$ in $S$ is true
(resp. false, known, unknown) in $\delta$.

The regression function ${\mathcal R}$ will be defined separately for
non-sensing actions and sensing actions to take into consideration
the fact that the execution of a non-sensing action
(resp. sensing action) in an a-state results in a single
a-state (resp. set of a-states). Thereafter, ${\mathcal R}$
 is extended
to define regression over conditional plans.
The key requirement on $\mathcal R$ is that it
should be sound (i.e., plans obtained through
regression must be plans based on the progression function) and complete
(i.e., for each plan based on progression, using regression one
should obtain that plan or a simpler plan with the same effects)
with respect to
progression.
We will also need to characterize the conditions
under which an action should not be used for regression.
Following \cite{bonet:geffner:hsp-01}, we refer to this condition
as ``the applicability
condition.''
We begin with non-sensing actions.

\subsection{Regression Over Non-Sensing Actions}

We begin with the
applicability condition of non-sensing actions and then give the
definition of the function ${\mathcal R}$ for non-sensing actions.
\begin{definition}
[Regression Applicability Condition -- Non-Sensing Action]
\label{app-non-sensing}
Given a non-sensing
action $a$ and a p-state $\delta=[ T,F ]$. We say that $a$ is
     {\em applicable} in $\delta$ if
\begin{enumerate}[(i)]
\item $Add_a \cap T \neq \emptyset$
     or $Del_a \cap F \neq \emptyset$, and
\item
     $Add_a \cap F = \emptyset$, $Del_a \cap T = \emptyset$,
     $Pre^+_a \cap F \subseteq Del_a$, and $Pre^-_a \cap T \subseteq Add_a$.
\end{enumerate}
\end{definition}
Intuitively, the aforementioned applicability condition requires
that $a$ is \emph{relevant} (item (i)) and \emph{consistent} (item
(ii)) in $\delta$. Item (i) is considered ``relevant'' as it makes
sure that the effects of $a$ will contribute to $\delta$ after
its execution. Item (ii) is considered ``consistent'' as it makes sure
that the situation obtained by progressing $a$, from a situation
yielded by regressing from $\delta$ through $a$, will be consistent with
$\delta$. Observe also that this definition will exclude the conventional
operator {\em no-op} from consideration for regression as it is never
applicable.

Since the application of
a non-sensing action in an a-state results in a single a-state,
the regression of a p-state over a non-sensing action should
result in a p-state.
This is defined next.
\begin{definition} [Regression -- Non-Sensing Action]
\label{reg-nonsensing} Given
a non-sensing action $a$ and a p-state $\delta = [T,F]$,
\begin{enumerate}[$\bullet$]
\item if $a$ is not applicable in
     $\delta$ then ${\mathcal R}(a,\delta) = \bot$;
\item if $a$ is applicable in $\delta$ then {${\mathcal R}(a,\delta) =
[ (T \setminus Add_a) \cup Pre_a^+, (F \setminus Del_a) \cup Pre_a^-
]$.}
\end{enumerate}
\end{definition}
Like in the progression function, the symbol $\bot$
indicates a ``failure.'' In other words, ${\mathcal R}(a,\delta) = \bot$
means that $\delta$ cannot be regressed on $a$ (or the regression
from $\delta$ over $a$ fails).
For later use, we extend the regression function ${\mathcal R}$ for
non-sensing actions over a set of p-states and define
  $${\mathcal R}(a,\{\delta_1, \ldots, \delta_n\}) =
\{{\mathcal R}(a,\delta_1), \ldots, {\mathcal R}(a,\delta_n)\}$$

  where $\delta_1, \ldots, \delta_n$ are p-states and $a$ is a
non-sensing action.

\begin{example} [Getting to Evanston -  con't] \label{ex4}

The actions $take\mbox{-}western$ and $take\mbox{-}ashland$ are applicable in
$\delta = [\{$at\mbox{-}evanston$\},\{\}]$.

  \hspace*{0.25in} ${\mathcal R}(take\mbox{-}western, \delta) =
[\{on\mbox{-}western\}, \{\mathit{traffic\mbox{-}bad}\}]$, and

  \hspace*{0.25in} ${\mathcal R}(take\mbox{-}ashland, \delta) =
[\{$on\mbox{-}ashland$\}, \{\}]$. 
\end{example}

\subsection{Regression Over Sensing Actions}

Let $a$ be a sensing action and $\sigma$ be an a-state. The
definition of the progression function $\Phi$ states that the
execution of $a$ in $\sigma$ results in a set of a-states
$\Phi(a,\sigma)$. Furthermore, if $a$ is executable in $\sigma$ then
every member of $\Phi(a,\sigma)$ extends $\sigma$ by a set of
fluents $s_a \subseteq Sens_a$ and every $f \in Sens_a \setminus
s_a$ is known in $\sigma$. As such, the regression over a sensing
action should be with respect to a set of p-states and result in a
p-state. Moreover, our definition for the applicability condition of
a sensing action must account for the fact that the set of p-states,
from which the regression is done, satisfies the two properties:
({\em i}) $Sens_a$ is known in each of its members; and ({\em ii})
the difference between two of its members is exactly $Sens_a$. This
leads to the following definition.

\begin{definition} [Sensed Set of Fluents]
\label{sensed-set-of-fluent} Let $\Delta =
\{\delta_1,\ldots,\delta_n\}$ be a set of p-states and $a$ be a
sensing action. A {\em sensed set of fluents} of $\Delta$ with
respect to $a$, denoted by $p(a,\Delta)$, is a non-empty subset of
$Sens_a$ satisfying the following properties:
\begin{enumerate}[$\bullet$]

\item $Sens_a$ is known in $\Delta$;
\item $n= 2^{|p(a,\Delta)|}$;
\item for every partition\footnote{
For a set of fluents $X$, a {\em partition} of $X$ is a pair of sets
of fluents $(P,Q)$ where $P \cap Q = \emptyset$ and $P \cup Q = X$.
} $(P,Q)$ of $p(a,\Delta)$, there exists only one $\delta_i \in
\Delta$ ($1 \leq i \leq n$) such that $\delta_i.T \cap p(a,\Delta) =
P,\ \delta_i.F \cap p(a,\Delta) = Q$; and
\item $\delta_i.T \setminus p(a,\Delta) = \delta_j.T \setminus
p(a,\Delta)$ and $\delta_i.F \setminus p(a,\Delta) = \delta_j.F \setminus p(a,\Delta)$
for every pair of $i$ and $j$, $1 \leq i \leq n$
and $1 \leq j \leq n$.
\end{enumerate}
\end{definition}

It can be seen from Definition \ref{def-tran} (Case 2) that
when a sensing action $a$ is executed in an a-state $\sigma$, the result
is a set of a-states $\Phi(a,\sigma)$ where for
each $\sigma' \in \Phi(a,\sigma)$, $\sigma' \setminus \sigma = Sens_a \setminus \sigma$.
The above definition captures the inverse of the progression process.
Intuitively,
$p(a,\Delta)$ is the set of fluents which are unknown
before the execution of $a$ and are known after its execution;
for example, if $\Delta = \Phi(a,\sigma)$, then $p(a,\Delta)$
should encode the set $Sens_a \setminus \sigma$.
It is easy to see that the second condition on $p(a,\Delta)$
warrants that it is a maximal subset of $Sens_a$ satisfying the
four stated conditions.
Observe also that due to the second condition, $\Delta$ must be
a non-empty set.
The next lemma proves that the sensed set of a set of p-states
with respect to an action is unique.

\begin{lemma} \label{lemSS-maintext}
For every sensing action $a$ and set of p-states $\Delta$,
$p(a,\Delta)$ is unique if it exists. 
\end{lemma}

\proof Abusing the notation, we write $p(a,\Delta) =
\bot$ whenever $p(a,\Delta)$ does not exist. Clearly, the lemma
holds if $\Delta = \emptyset$ as $p(a,\Delta) = \bot$ for every $a$.
So, we need to prove it for the case $\Delta \ne \emptyset$.

Assume that $p(a,\Delta)$ exists but it is not unique, i.e, we can
find different sensed sets of $\Delta$ with respect to $a$, say $X$
and $X'$. By Definition \ref{sensed-set-of-fluent}, we have that $X
\neq \emptyset$ and $X' \neq \emptyset$.

Since $X \neq \emptyset$, let us consider a fluent $f \in X$. By
Definition \ref{sensed-set-of-fluent}, for two partitions $(\{f\}, X
\setminus \{f\})$ and $(X \setminus \{f\}, \{f\})$ of $X$, there
exist $\delta \in \Delta$ and $\delta' \in \Delta$ such that $\{f\}
= \delta.T \cap X$ and $\{f\} = \delta'.F \cap X$, i.e. $f$ is true
in $\delta$ and false in $\delta'$. \hfill[*].

Suppose that $f \not \in X'$. By Item 4, Definition
\ref{sensed-set-of-fluent}, either $f \in \delta.T \setminus X'$ or
$f \in \delta.F \setminus X'$ for $\delta \in \Delta$, i.e $f$ is
either true or false in every $\delta \in \Delta$. In either case,
this contradicts with [*]. Therefore, $f \in X'$.

Symmetrically, we can argue that, if $f \in X'$ then $f \in X$.
Thus, $f \in X$ iff $f \in X'$, i.e. $X=X'$.\qed
%
\begin{definition} [Properness] \label{sensed-Set}
A set of p-states $\Delta$
is {\em proper} with respect to a sensing action $a$
if $p(a,\Delta) \ne \bot$.
\end{definition}
For convenience, we sometime write $p(a,\Delta) = \bot$ to indicate that
$\Delta$ is not proper with respect to $a$.
\begin{example} [Getting to Evanston --  Cond't] \label{ex41} Consider a
set $\Delta_1 = \{ \delta_1, \delta_2\}$ where $\delta_1 = [
\{at\mbox{-}start, \mathit{traffic\mbox{-}bad}\},
\{$on\mbox{-}western, on\mbox{-}belmont,
on\mbox{-}ashland, at\mbox{-}evanston$\} ]$ and $$\delta_2 = [
\{at\mbox{-}start\}, \{\mathit{traffic\mbox{-}bad},
on\mbox{-}western, on\mbox{-}belmont,
on\mbox{-}ashland, at\mbox{-}evanston\} ].$$
We can easily check that
\[
p(\mathit{check\mbox{-}traffic},\Delta_1) = \{\mathit{traffic\mbox{-}bad}\}.
\]
So, $\Delta_1$ is proper with respect to
$check\mbox{-}\mathit{traffic}$.

On the other hand
\[
p(\mathit{check\mbox{-}traffic},\Delta_2) = \bot.
\]
where $\Delta_2 = \{\delta_1, \delta_3\}$ and $\delta_3 =
[ \{at\mbox{-}start\}, \{\mathit{traffic\mbox{-}bad}, at\mbox{-}evanston\}]$.
This is because the fourth condition (Definition \ref{sensed-set-of-fluent})
cannot be satisfied for any non-empty subset of $\mathit{traffic\mbox{-}bad}$.
So, $\Delta_2$ is not proper with respect to
$check$-$\mathit{traffic}$.
\end{example}

We are now ready to define the applicability condition for
sensing actions. The definition is given in two steps.
First, we define the strong applicability condition as follows.

\begin{definition} [Strong Regression Applicability Condition -- Sensing Action]
\label{app-sensing-strong}  Let $a$ be a sensing action and
$\Delta$ be a set of p-states. We
say that $a$ is \emph{strongly applicable} in $\Delta$ if
\begin{enumerate}[(i)]
\item
$p(a,\Delta) \neq \bot$; and
\item $Pre^+_a \cap \delta.F =
\emptyset$ and $Pre^-_a \cap \delta.T = \emptyset$
for every $\delta \in \Delta$.
\end{enumerate}
\end{definition}
In the above definition, (i) and (ii) correspond to the ``relevancy''
and ``consistency'' requirement for non-sensing actions (Definition
\ref{app-non-sensing}) respectively.
(i) corresponds to the fact that
executing a sensing action $a$ in an a-state $\sigma$ results in a
set of $2^{|p(a,\Delta)|}$ a-states, each of which extends $\sigma$
by $p(a,\Delta)$  and (ii) guarantees that $a$ must be executable
prior to its execution.

  It is easy to see that if a sensing action $a$ is strongly
applicable in $\Delta$, then for every a-state $\sigma'$ extending
the p-state
$$\delta' =
     [ ((\bigcup_{\delta \in \Delta}\delta.T) \setminus p(a,\Delta)) \cup Pre^+_a,
       ((\bigcup_{\delta \in \Delta}\delta.F) \setminus p(a,\Delta)) \cup Pre^-_a ]
$$
it holds that every member of $\Phi(a, \sigma')$ belongs to the
extension of some $\delta_i \in \Delta$. As such, $\delta'$ could
be viewed as the result of the regression from $\Delta$ through $a$.
Unfortunately, the conditions in Definition
\ref{app-sensing-strong} are sometime unnecessarily strong as the
following example demonstrates.

\begin{example}
[Strong Regression Applicability Condition]
\label{ex41strong}
Let $$P = \langle \{f,g,h\}, \{sense_f, a_1, a_2\}, \{h\},
\{g\} \rangle$$ be a planning problem, where $sense_f$ is a sensing action with
$$Pre_{sense_f} = \{h\}, \; Sens_{sense_f} = \{f\};$$ $a_1$ and
$a_2$ are two non-sensing actions with
$$Pre_{a_1} = \{h,f\}, \; Add_{a_1} = \{g\}, Del_{a_1} = \emptyset,$$
$$Pre_{a_2} = \{\neg f\}, \; Add_{a_2} = \{g\}, \textnormal{
and } Del_{a_2} = \emptyset.$$

Clearly,
$$c = sense_f; case(f \rightarrow a_1,  \neg f \rightarrow a_2)$$
is a progression solution to $P$. Thus, it is reasonable to expect
that if we regress from the goal $\delta = [\{g\}, \emptyset]$ on
$c$ --- step-by-step --- we will receive a p-state $\delta'$ such that
$\langle \{h\}, \emptyset \rangle \in ext(\delta')$.
This process begins with the
regression on $a_1$ and $a_2$ from $\delta$. Thereafter, we
receive a set of p-states from which the regression
on $sense_f$ can be done. It is easy to see that ${\mathcal R}(a_1, \delta) =
[\{h,f\}, \emptyset] = \delta_1$ and ${\mathcal R}(a_2, \delta) =
[\emptyset, \{f\}] = \delta_2$. It is also easy to see that the
strong applicability condition implies that $sense_f$ is not
applicable in $\{\delta_1 , \delta_2\}$ because $p(sense_f,
\{\delta_1 , \delta_2\}) = \bot$. This means that, we cannot
regress on $c$ from the goal state. 
\end{example}

Notice that the problem in the above example lies in
the fact that $\{\delta_1, \delta_2\}$ violates the properness
definition in that $\delta_1$ and $\delta_2$
do not have the same values
on the set of fluents that do not belong to $Sens_{sense_f}$.
To overcome the problem posed by the strong applicability condition,
we relax this condition.

\begin{definition}
[Regression Applicability Condition -- Sensing Action]
\label{app-sensing}
Let $a$ be a sensing action and $\Delta$ be a set of p-states. $a$ is
\emph{applicable} in $\Delta$ if
\begin{enumerate}[(i)]
\item  there exists a set of p-states $\Delta'$ such that
$\Delta \sqsubseteq \Delta'$ and $a$
is strongly applicable in $\Delta'$; and
\item  $Sens_a$ is known in $\Delta$.
\end{enumerate}
\end{definition}

\begin{example} [Continuation of Example \ref{ex41strong}]
\label{ex41strong-add}
It is easy to see that $sense_f$ is applicable in $\{\delta_1,\delta_2\}$
since it is strongly applicable in $\{\delta_1,\delta_2'\}$
where $\delta_2' = [\{h\},\{f\}]$ and
$\{\delta_1,\delta_2\} \sqsubseteq \{\delta_1,\delta_2'\}$.
\end{example}

\begin{definition}
[Sensed Set]
Let $\Delta$ be a set of p-states and $a$ be a sensing action
such that $a$ is applicable in $\Delta$. We say that $X$ is
a sensed set of fluents of $\Delta$ with respect to $a$,
denoted by $S_{a,\Delta}$,
if there exists a set of p-states $\Delta'$ such that
$\Delta \sqsubseteq \Delta'$,
$a$ is strongly applicable in $\Delta'$, and
$X = p(a,\Delta')$.
\end{definition}
Again, we write $S_{a,\Delta} = \bot$ to say that
the sensed set of fluents of $\Delta$ with respect to $a$
does not exist.
The next lemma states that $S_{a,\Delta}$ is unique.

\begin{lemma} \label{lemSS1-maintext}
For every sensing action $a$ and set of p-states $\Delta$,
$S_{a,\Delta}$ is unique if it exists. 
\end{lemma}

\proof Obviously, the lemma holds for
$\Delta=\emptyset$. So, we need to prove it for the case $\Delta \ne
\emptyset$.

Assume the contrary, $S_{a,\Delta}$ is not unique. This implies that
there exists $\Delta'$ and $\Delta''$ such that $\Delta \sqsubseteq
\Delta'$ and $\Delta \sqsubseteq \Delta''$, $p(a,\Delta') \neq
\bot$, $p(a,\Delta'') \neq \bot$, and $p(a,\Delta') \neq
p(a,\Delta'')$. Again, by Definition \ref{sensed-set-of-fluent} we
can conclude that $p(a,\Delta') \neq \emptyset$ and $p(a,\Delta'')
\ne \emptyset$. Without loss of generality, we conclude that there
exists some $f \in p(a,\Delta') \setminus p(a,\Delta'')$.

Since $p(a,\Delta'') \ne \bot$, $f \in Sens_a$, and $f$ is known in
$\Delta''$, by Definition \ref{sensed-set-of-fluent}, we must have
two cases.

\begin{enumerate}[(1)]
\item $f \in
{\delta}''.T \setminus p(a,\Delta'')$ for every $\delta'' \in
\Delta''$. Since $f \in p(a,\Delta')$, by Definition
\ref{sensed-set-of-fluent}, for the partition $(p(a,\Delta')
\setminus \{f\}, \{f\})$ of $p(a,\Delta')$, there exists $\delta'
\in \Delta'$ such that $\delta'.F \cap p(a,\Delta') = \{f\}$, i.e.
$f$ is false in $\delta'$.

Because $\Delta \sqsubseteq \Delta'$, there exists some $\delta \in
\Delta$ such that $\delta'$ is a partial extension of $\delta$. So,
 we have that $\delta.F \subseteq \delta'.F$.
Also, as $Sens_a$ is known in $\delta$, we must have that $f \in
\delta.F$.

Since $\Delta \sqsubseteq \Delta''$, we know that there exists some
${\delta}'' \in \Delta''$ which is a partial extension of $\delta$.
This implies that $\delta.F \subseteq {\delta}''.F$, i.e., $f \in
{\delta}''.F$. This contradicts with the fact that $f \in
{\delta}''.T$.

\item $f \in {\delta}''.F \setminus p(a,\Delta'')$ for
every $\delta'' \in \Delta''$. Similarly to the first case, we can
derive a contradiction.
\end{enumerate}
The above two cases show that if $f \in p(a,\Delta')$ then $f \in
p(a,\Delta'')$.

This shows that
$p(a,\Delta') = p(a,\Delta'')$.\qed 

We illustrate the above definition in the next example.

\begin{example} [Getting to Evanston -  con't] \label{ex42}
Consider the
set $\Delta_2$ and the sensing action $\mathit{check\mbox{-}traffic}$ in Example
\ref{ex41}. We have that

\begin{enumerate}[(i)]
     \item $\mathit{check\mbox{-}traffic}$ is not strongly applicable in $\Delta_2$
        (because $p(\mathit{check\mbox{-}traffic},\Delta_2) = \bot$, Example \ref{ex41});
        however,
     \item  $\mathit{check\mbox{-}traffic}$ is applicable in $\Delta_2$.
This is because $\Delta_1$ (Example \ref{ex41}) consists of partial
extensions of p-states in $\Delta_2$, and
$\mathit{check\mbox{-}traffic}$ is
strongly applicable in $\Delta_1$.
\end{enumerate}
\end{example}
We are now ready to define the regression function for sensing actions.

\begin{definition} [Regression -- Sensing Action] \label{reg-sensing}
Let $a$ be a sensing action and $\Delta$ be a set of p-states.
\begin{enumerate}[$\bullet$]
   \item  if $a$ is not applicable in
     $\Delta$ then ${\mathcal R}(a, \Delta) = \bot$; and
   \item  if $a$ is applicable in
     $\Delta$
\[
{\mathcal R} (a, \Delta) =
     [ ((\bigcup_{\delta \in \Delta} \delta.T) \setminus S_{a,\Delta}) \cup Pre^+_a,
       ((\bigcup_{\delta \in \Delta} \delta.F) \setminus S_{a,\Delta}) \cup Pre^-_a ].
\]
\end{enumerate}
\end{definition}

\begin{example} [Getting to Evanston --  Cond't] \label{ex5}
The action $check\mbox{-}\textnormal{\em traffic}$ is applicable in
$\Delta_2$ with respect to $\{\textnormal{\em traffic}\mbox{-}bad\}$
(see Example \ref{ex42}) and we have

\hspace{.25in} ${\mathcal R}(check\mbox{-}\textnormal{\em traffic}, \Delta_2) =$

\hspace{.65in} $[\{at\mbox{-}start\},\{on\mbox{-}western,
on\mbox{-}belmont, on\mbox{-}ashland, at\mbox{-}evanston\} ].$
\end{example}

\subsection{Regression Over Conditional Plans}

We now extend
${\mathcal R}$ to define ${\mathcal R}^*$ that allows us to perform
regression over conditional plans. For a conjunction of fluent
literals, by $\varphi^+$ and
$\varphi^-$ we denote the sets of fluents occurring positively
and negatively in $\varphi$, respectively.

\begin{definition}[Extended Regression Function] \label{d3}
Let $\delta$ be a p-state. The extended transition function
${\mathcal R}^*$ is defined as follows:

\begin{enumerate}[$\bullet$]

\item  ${\mathcal R}^*([\ ], \delta) = \delta$.

\item  For a non-sensing action $a$, ${\mathcal R}^*(a,
\delta)$ = ${\mathcal R}(a, \delta)$.

\item  For a conditional plan $p  = a;case(\varphi_1
{\rightarrow} c_1, \ldots, \varphi_n {\rightarrow} c_n)$
where $a$ is a sensing action and $c_i$'s are conditional
plans,
\begin{enumerate}[--]

\item  if ${\mathcal R}^*(c_i, \delta) {=} \bot$ for some $i$, $
{\mathcal R}^*(p,\delta) = \bot$;

\item if ${\mathcal R}^*(c_i, \delta) {=} [T_i,F_i]$
for $i=1,\ldots,n$, then
\[
{\mathcal R} ^*(p,\delta) = {\mathcal R}(a,\{ R(\varphi_1 \rightarrow c_1,\delta),
\ldots, R(\varphi_n \rightarrow c_n,\delta)\})\]
 where $R(\varphi_i \rightarrow c_i, \delta) = [T_i \cup
\varphi_i^+,F_i \cup \varphi_i^-]$ if $\varphi_i^+ \cap F_i =
\emptyset$ and $\varphi_i^- \cap T_i = \emptyset$; otherwise,
$R(\varphi_i \rightarrow c_i, \delta) = \bot$.
\end{enumerate}

\item  For $p = a; c$, where $a$ is a non-sensing action and $c$ is
a conditional plan,
\[
{\mathcal R}^*(p,\delta) = {\mathcal R}(a,{\mathcal R}^* (c, \delta));
\]
\item  ${\mathcal R}^* (p,\perp) = \perp$ for
every plan $p$.
\end{enumerate}
\end{definition}
The notion of a regression solution is defined as follows.
\begin{definition}
[Regression Solution] \label{reg-sol} A conditional plan $c$ is a
{\em regression solution} to the planning problem $P =\langle
A,O,I,G \rangle$ if ${\mathcal R}^*(c, \delta_G) \ne \bot$ and
$\sigma_I \in ext({\mathcal R}^*(c, \delta_G))$ where $\delta_G =
[G^+, G^- ]$ and $\sigma_I = \langle I^+, I^- \rangle$.
\end{definition}
The above definition is a generalization of the notion of a plan
obtained by regression in domains without sensing actions and
with complete information about the initial state to domains with
sensing actions and incomplete information. An important property
that any regression function needs to satisfy is its soundness with
respect to the corresponding progression function. Here, we would
like to guarantee that $\mathcal{R}$ and $\mathcal{R}^*$ are sound
with respect to the progression function $\Phi$ and $\Phi^*$, respectively. As
such, we require that a regression solution $c$ to a planning
problem $P =\langle A,O,I,G \rangle$ be a plan achieving the goal
$G$ from $I$. This property is proved in Theorem
\ref{lem3-maintext}.

As the soundness of the regression function with respect to the
progression function is guaranteed, it will be interesting to
investigate its completeness. In this paper, we opt for a definition
that --- when used in planning --- will give us optimal solutions in
the sense that regression solutions do not contain redundant
actions. This is evident from Definitions \ref{app-non-sensing} and
\ref{app-sensing} in which we require that the action, over which
the regression is done, must add new information to the regressed
state. We will elaborate in more detail on this point in the next
section.

\section{Soundness and Completeness Results}
\label{result}

In this section, we show that our regression function ${\mathcal R}^*$
is sound and complete with respect to the progression function
$\Phi$. 

\subsection{Soundness Result}

As with its definition, the soundness of ${\mathcal R}^*$
is proved in two steps. First, we prove the soundness of $\mathcal R$, 
separately for non-sensing and sensing actions. 
Second, we extend this result to regression solutions. 
To establish the soundness of ${\mathcal R}$ 
on non-sensing actions, we need the following lemma.

{\lemma \label{lem1-appdx} Let $\delta$ be a p-state. An a-state
$\sigma$ is an extension of $\delta$ (i.e. $\sigma \in ext(\delta)$)
iff $\sigma$ is an a-state of the form $\langle \delta.T \cup X,
\delta.F \cup Y \rangle$ where $X, Y$ are two disjoint sets of
fluents and $X \cap \delta.F = \emptyset,\ Y \cap \delta.T =
\emptyset$.  
}

\proof\hfill
\begin{enumerate}[$\bullet$]
     \item  Case ``$\Rightarrow$'':

     Let $\sigma \in ext(\delta)$ be an extension of $\delta$. By
     the definition of an extension, $\sigma$ is an a-state where $\delta.T \subseteq
     \sigma.T$ and $\delta.F \subseteq \sigma.F$. Denote $X=\sigma.T \setminus
     \delta.T$ and $Y = \sigma.F \setminus \delta.F$. Clearly, $X$
     and $Y$ are two set of fluents where $X \cap Y = \emptyset$ and
     $X \cap \delta.F = \emptyset,\ Y \cap \delta.T = \emptyset$.
     \item  Case ``$\Leftarrow$'':

     Let $\sigma$ be an a-state of the form $\langle \delta.T \cup
     X, \delta.F \cup Y \rangle$ where $X, Y$ are two disjoint
     sets of fluents and $X \cap \delta.F = \emptyset,\ Y \cap \delta.T
     = \emptyset$.

     It's easy to see that $\sigma.T \cap \sigma.F = \emptyset$,
     i.e. $\sigma$ is consistent. Furthermore, $\delta.T \subseteq
     \sigma.T$ and $\delta.F \subseteq \sigma.F$, i.e. by definition
     of an extension, $\sigma$ is an extension of $\delta$.\qed
\end{enumerate}

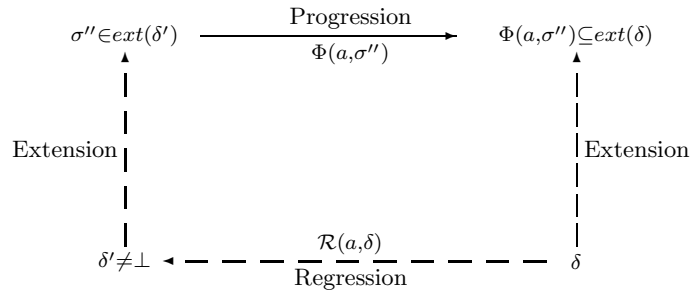
\begin{figure}[htb]
\begin{center}
\setlength{\unitlength}{1cm}
\begin{picture}(6,3.5)
\multiput(5.6,0)(-0.5,0){10}{\line(-1,0){0.3}}
\put(0.5,0){\vector(-1,0){0}}
\put(6,0){\makebox(0,0)[cc]{$_\delta$}}

\put(3,0.1){\makebox(0,0)[cb]{$_{{\mathcal R}(a,\delta)}$}}

\put(3,2.9){\makebox(0,0)[ct]{$_{\Phi(a,\sigma'')}$}}

\multiput(0,0.2)(0,0.5){5}{\line(0,1){0.3}}
\put(0,2.8){\vector(0,1){0}}
\put(0,0){\makebox(0,0)[cc]{$_{\delta' \ne \bot}$}}

\put(0,3){\makebox(0,0)[cc]{$_{\sigma'' \in ext(\delta')}$}}

\put(3,3.1){\makebox(0,0)[cb]{\footnotesize Progression}}

\put(6.1,1.5){\makebox(0,0)[lc]{\footnotesize Extension}}
\put(-0.1,1.5){\makebox(0,0)[rc]{\footnotesize Extension}}
\put(3,-0.1){\makebox(0,0)[ct]{\footnotesize Regression}}

\put(1,3){\vector(1,0){3.4}}

\multiput(6,0.2)(0,0.4){6}{\line(0,1){0.3}}
\put(6,2.8){\vector(0,1){0}}

\put(6,3){\makebox(0,0)[cc]{$_{\Phi(a,\sigma'') \subseteq
ext(\delta)}$}}
\end{picture}
     \caption{Illustration of Theorem \ref{lem1-maintext}.}
     \label{example-fig-02}
\end{center}
\end{figure}

Intuitively, the soundness of ${\mathcal R}$ for a non-sensing action 
 states that the regression over a
non-sensing action from a p-state yields another p-state such that
the execution of the action in any extension of the latter results
in a subset of a-states belonging to the extension set of the
former. This is illustrated in Figure
\ref{example-fig-02}.

\begin{theorem} [Non-sensing Action] \label{lem1-maintext}
Let $\delta$ be a p-state and $a$ be a non-sensing action. If
${\mathcal R}(a,\delta)= \delta'$ and $\delta' \neq \bot$, then for
every $\sigma'' \in ext(\delta')$ we have that $\Phi(a, \sigma'')
\subseteq ext(\delta)$.   
\end{theorem}

\proof Let $\delta =
[T,F]$. From the fact that ${\mathcal R}(a,\delta)= \delta' \neq
\bot$, we have that $a$ is applicable in $\delta$.

\st By Definition \ref{reg-nonsensing},
\[
\delta' = {\mathcal R}(\delta,a) = [(T \setminus Add_a) \cup
Pre^+_a, (F \setminus Del_a) \cup Pre^-_a].
\]

Let $\sigma'' \in ext(\delta')$. It follows from Lemma
\ref{lem1-appdx} that
\[
\sigma'' = \langle (T \setminus Add_a) \cup Pre^+_a \cup X, (F
\setminus Del_a) \cup Pre^-_a \cup Y \rangle,
\]
where $X$ and $Y$ are two sets of fluents such that $\sigma''.T \cap
\sigma''.F = \emptyset$. We now prove that (i) $a$ is executable in
$\sigma''$ and (ii) $\Phi(a, \sigma'') \subseteq ext(\delta)$.

\begin{enumerate}[$\bullet$]
     \item  Proof of (i):

Since $Pre^+_a \subseteq \sigma''.T$ and $Pre^-_a \subseteq
\sigma''.F$, we conclude that ${lem1-maintext}a$ is executable in $\sigma''$.

     \item  Proof of (ii):

\st By definition of the transition function $\Phi$, we have that
\[
\Phi(a, \sigma'') = \{\langle (((T \setminus Add_a) \cup Pre^+_a
\cup X) \setminus Del_a) \cup Add_a, (((F \setminus Del_a) \cup
Pre^-_a \cup Y) \setminus Add_a) \cup Del_a \rangle\}
\]

\st Since $a$ is applicable in $\delta$, we have that $T \cap Del_a
= \emptyset$, $F \cap Add_a = \emptyset$. Furthermore, $Del_a \cap
Add_a = \emptyset$. Therefore, we have that $(((T \setminus Add_a)
\cup Pre^+_a \cup X) \setminus Del_a) \cup Add_a = ((T \setminus
Add_a) \cup ((Pre^+_a \cup X) \setminus Del_a)) \cup Add_a \supseteq
T \cup ((Pre^+_a \cup X) \setminus Del_a) \supseteq T$. This
concludes that $T \subseteq \Phi(a, \sigma'').T$. Similarly, we have
that $F \subseteq \Phi(a, \sigma'').F$. This shows that $\Phi(a,
\sigma'') \subseteq ext(\delta)$.\qed
\end{enumerate}

Observe that the conclusion of the theorem indicates that
$\bot \not \in \Phi(a, \sigma'')$, i.e., $a$ is executable in
$\sigma''$. This shows that ${\mathcal R}$ can be ``reversed'' for
non-sensing actions. 

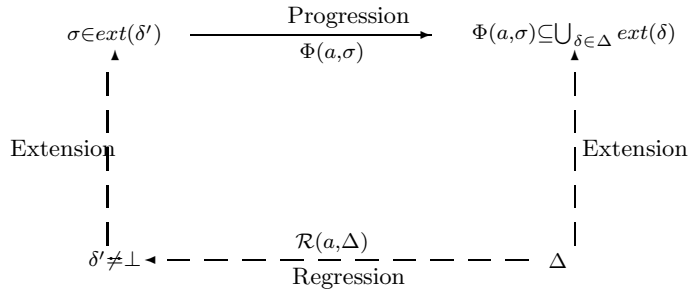
\begin{figure}[htb]
\begin{center}
\setlength{\unitlength}{1cm}
\begin{picture}(6,3.5)
\multiput(5.6,0)(-0.5,0){10}{\line(-1,0){0.3}}
\put(0.5,0){\vector(-1,0){0}}
\put(6,0){\makebox(0,0)[cc]{$_\Delta$}}

\put(3,0.1){\makebox(0,0)[cb]{$_{{\mathcal R}(a,\Delta)}$}}

\put(3,2.9){\makebox(0,0)[ct]{$_{\Phi(a,\sigma)}$}}

\multiput(0,0.2)(0,0.5){5}{\line(0,1){0.3}}. 
\put(0,2.8){\vector(0,1){0}}
\put(0,0){\makebox(0,0)[cc]{$_{\delta' \ne \bot}$}}

\put(0,3){\makebox(0,0)[cc]{$_{\sigma \in ext(\delta')}$}}

\put(1,3){\vector(1,0){3.2}}. 

\multiput(6,0.2)(0,0.5){5}{\line(0,1){0.3}}
\put(6,2.8){\vector(0,1){0}}

\put(6,3){\makebox(0,0)[cc]{$_{\Phi(a,\sigma) \subseteq
\bigcup_{\delta \in \Delta} ext(\delta)}$}}

\put(3,3.1){\makebox(0,0)[cb]{\footnotesize Progression}}
\put(6.1,1.5){\makebox(0,0)[lc]{\footnotesize Extension}}
\put(-0.1,1.5){\makebox(0,0)[rc]{\footnotesize Extension}}
\put(3,-0.1){\makebox(0,0)[ct]{\footnotesize Regression}}
\end{picture}
\end{center}
     \caption{Illustration of Theorem \ref{lem2-maintext}.}
     \label{example-fig-03}. 
\end{figure}

We will next establish a result similar to Theorem \ref{lem1-maintext}
for sensing actions.  Intuitively, the result should state that
the regression over a 
sensing action from a set of p-states yields a p-state such that the
execution of the action in any extension of the latter results in
a set of a-states belonging to the union of the extension sets
of the former, i.e., it should allow us to conclude that ${\mathcal R}$ can be ``reversed''
for sensing actions. Figure \ref{example-fig-03} illustrates this idea.

We need the following lemma.

{\lemma \label{lem0-appdx} Let $\sigma'$ be an a-state and $a$ be a
sensing action executable in $\sigma'$. For any $S_a \subseteq
Sens_a$ and $\sigma \in \Phi(a,\sigma')$, let $\sigma.T \cap S_a =
S^+_\sigma$ and $\sigma.F \cap S_a = S^-_\sigma$, we have that
$S^+_\sigma \cup S^-_\sigma = S_a$ and $S^+_\sigma \cap S^-_\sigma =
\emptyset$.  
}

\proof It is easy to see that the lemma is correct
for the case $S_a = \emptyset$. Let us consider the case $S_a \neq
\emptyset$. Since $S^+_\sigma \subseteq \sigma.T$ and $S^-_\sigma
\subseteq \sigma.F$, we have that $S^+_\sigma \cap S^-_\sigma =
\emptyset$.
. This also shows
Consider $f \in S^+_\sigma \cup S^-_\sigma$, we have that $f \in
S^+_\sigma$ or $f \in S^-_\sigma$. In both cases, we have $f \in
S_a$.

Consider $f \in S_a$. Since $S_a \subseteq Sens_a$, we have that $f
\in Sens_a$. By the definition of $\Phi$, we have that $f \in
\sigma.T$ or $f \in \sigma.F$. From this fact, it's easy to see
that $f \in S^+_\sigma$ or $f \in S^-_\sigma$.\qed 

With the help of the above lemma, we can prove the following theorem. 

\begin{theorem} [Sensing action] \label{lem2-maintext} Let
$\Delta$ be a set of p-states and $a$ be a sensing action. If
${\mathcal R}(a,\Delta) = \delta'$ and $\delta' \neq \bot$, then for
every $\sigma'' \in ext(\delta')$, we have that $\Phi(a,\sigma'')
\subseteq \bigcup_{\delta \in \Delta} ext(\delta)$. 
\end{theorem}

\proof From the fact
that ${\mathcal R}(a,\Delta) = \delta' \neq \bot$, we have that $a$
is applicable in $\Delta$ with respect to some set $S_{a,\Delta}
\subseteq Sens_a$ ($S_{a,\Delta} \neq \emptyset$). By Definition
\ref{reg-sensing} we have:

\[ \delta' = {\mathcal R}(a,\Delta) = [
(\bigcup_{\delta \in \Delta}\delta.T \setminus S_{a,\Delta}) \cup
Pre^+_a, (\bigcup_{\delta \in \Delta}\delta.F \setminus
S_{a,\Delta}) \cup Pre^-_a ].
\]
Let $\sigma'' \in ext(\delta')$ be an arbitrary extension of
$\delta'$. We will now prove (i) $a$ is executable in $\sigma''$ and
(ii) $\Phi(a,\sigma'') \subseteq \bigcup_{\delta \in \Delta}
ext(\delta)$.

\begin{enumerate}[$\bullet$]
     \item Proof of (i):
     It follows from Lemma \ref{lem1-appdx} that
\[
\sigma'' = \langle (\bigcup_{\delta \in \Delta}\delta.T \setminus
S_{a,\Delta}) \cup Pre^+_a \cup X, (\bigcup_{\delta \in
\Delta}\delta.F \setminus S_{a,\Delta}) \cup Pre^-_a \cup Y \rangle
\]

\st where $X$ and $Y$ are two sets of fluents such that $\sigma''.T
\cap \sigma''.F = \emptyset$.

\st From the fact that $Pre^+_a \subseteq \sigma''.T$ and
  $Pre^-_a \subseteq \sigma''.F$, we conclude that $a$ is executable in
$\sigma''$. \hfill[*]

\item Proof of (ii):
Consider an arbitrary $\sigma \in \Phi(a,\sigma'')$. We need to
prove that there exists some $\delta \in \Delta$ such that $\sigma
\in ext(\delta)$.

Since $a$ is applicable in $\Delta$, by Definition
\ref{app-sensing}, there exists $\Delta'$ such that $\Delta
\sqsubseteq \Delta'$ and $a$ is strongly applicable in $\Delta'$ and
$S_{a,\Delta} = p(a,\Delta')$.

Let $S^+_\sigma = \sigma.T \cap S_{a,\Delta}$ and $S^-_\sigma =
\sigma.F \cap S_{a,\Delta}$. By Lemma \ref{lem0-appdx}, we have that
$S^+_\sigma \cup S^-_\sigma = S_{a,\Delta}$ and $S^+_\sigma \cap
S^-_\sigma = \emptyset$.

By Definition \ref{sensed-set-of-fluent}, there exists $\delta' \in
\Delta'$ such that $\delta'.T \cap S_{a,\Delta} = S^+_\sigma$ and
$\delta'.F \cap S_{a,\Delta} = S^-_\sigma$. Because $\Delta
\sqsubseteq \Delta'$, there exists some $\delta \in \Delta$ such
that $\delta'$ is a partial extension of $\delta$.

We will show that $\sigma \in ext(\delta)$, i.e., $\delta.T
\subseteq \sigma.T$ and $\delta.F \subseteq \sigma.F$.

Since $\delta.T \subseteq \delta'.T$, we have that $\delta.T \cap
S_{a,\Delta} \subseteq \delta'.T \cap S_{a,\Delta} = S^+_\sigma$.
Therefore:
\[
\begin{array}{lll}
\delta.T & = & (\delta.T \setminus (\delta.T \cap S_{a,\Delta}))
\cup
(\delta.T \cap S_{a,\Delta}) \\
 & = & (\delta.T \setminus
S_{a,\Delta}) \cup ( \delta.T \cap S_{a,\Delta}) \\
& \subseteq & (\delta.T \setminus S_{a,\Delta}) \cup S^+_\sigma.
\end{array}
\]

Similarly, we can show that $\delta.F \subseteq (\delta.F \setminus
S_{a,\Delta}) \cup S^-_\sigma$.

Since $\sigma \in \Phi(a, \sigma'')$, by the definition of $\Phi$,
we have that $\sigma''.T \subseteq \sigma.T$. Let $\sigma.T
\setminus \sigma''.T = \omega$, we have that
\[\sigma.T =
\sigma''.T \cup \omega = ((\bigcup_{\delta \in \Delta}\delta.T)
\setminus
                               S_{a,\Delta}) \cup Pre^+_a \cup X \cup \omega.
\]
Since $\sigma.T \cap S_{a,\Delta} = S^+_\sigma$ and
$(((\bigcup_{\delta \in \Delta}\delta.T) \setminus S_{a,\Delta})
\cup Pre^+_a) \cap S_{a,\Delta} = \emptyset$ (because $Sens_a \cap
Pre^+_a = \emptyset$), we must have that $(X \cup \omega) \cap
S_{a,\Delta} = S^+_\sigma$, i.e. $S^+_\sigma \subseteq X \cup
\omega$. From the fact that $\delta.T \subseteq (\delta.T \setminus
S_{a,\Delta}) \cup S^+_\sigma$ and $S^+_\sigma \subseteq X \cup
\omega$, it is easy to see that $\delta.T \subseteq \sigma.T$.
Similarly, we can show that $\delta.F \subseteq \sigma.F$. From this
fact, we conclude that $\sigma \in ext(\delta)$. \hfill[**]

\end{enumerate}

From [*] and [**] the theorem is proved.\qed 

\bigskip
To prove the final result about the correctness of ${\mathcal R}$
(and Theorem \ref{theoAdd1-maintext} in the next section) we need a number of
additional notations and definitions.

\begin{definition} [Branching Count] Let $c$ be a conditional plan,
we define the number of case plans of $c$, denoted by $count(c)$,
inductively as follows:
\begin{enumerate}[(1)]
     \item if $c=[\ ]$ then $count(c) = 0$;
     \item if $c = a$, $a$ is a non-sensing action, then $count(c) = 0$;
     \item if $a$ is a non-sensing action and $c$ is a conditional plan
then $count(a;c) = count(c);$
     \item if $c$ is a case plan of the form
         a; case($\varphi_1 \rightarrow c_1, \ldots, \varphi_n \rightarrow c_n $)
         where $a$ is a sensing action, then
         $count(c) = 1 + \sum^n_{i=1} count(c_i).$
\end{enumerate}
\end{definition}

{\lemma [Sequence of Non-sensing Action] \label{cor1} For p-states
$\delta$ and $\delta'$, and a sequence of non-sensing actions $c =
a_1; \ldots; a_n$ $(n \geq 0)$, ${\mathcal R}^*(c,\delta)= \delta'
\neq \bot$ implies that $\Phi^*(c,\sigma'') \subseteq ext(\delta)$
for every $\sigma'' \in ext(\delta')$. 
}

\proof By induction on $n$.

\begin{enumerate}[$\bullet$]
     \item  Base Case: $n = 1$.
         This means that $c$ has only one action $a$. Using
         Theorem \ref{lem1-maintext}, and Definition \ref{d3} -- item 2 -- the base case is
         proved. Notice that for the case $n = 0$, i.e. $c=[\ ]$,
         the lemma follows directly from Definitions \ref{d3}
         and \ref{def-extran}.
     \item  Inductive Step:

         Assume that the lemma is shown for $1 \le n \leq k$.
         We now prove the lemma for $n = k+1$.

         Let $c'=a_2;\ldots; a_{k+1}$ and  ${\mathcal R}^*(c',\delta) = \delta^*$.
         By Definition \ref{d3}

         \[
         {\mathcal R}^*(c,\delta) = {\mathcal R}(a_1, {\mathcal R}^*(c',\delta)) = \delta'.
         \]

         \st Since ${\mathcal R}(a_1, \delta^*) = \delta' \neq \bot$,
         we have that $\delta^* \neq \bot$.

         \st Let $\sigma'' \in ext(\delta')$.
    By Theorem \ref{lem1-maintext}, we have
    that
    $\Phi(a_1,\sigma'') = \{\sigma\} \subseteq
         ext(\delta^*)$, i.e.,  $\sigma \in ext(\delta^*)$.

         \st By the definition of $\Phi^*$, we also have
    that $\Phi^*(c, \sigma'') = \Phi^*(c', \Phi^*(a_1,\sigma''))$.
         Using the induction hypothesis for $c'$,
    where ${\mathcal R}^*(c',\delta) = \delta^*$ and $\sigma \in ext(\delta^*)$, we have:
         \[
    \Phi^*(c', \Phi^*(a_1, \sigma'')) = \Phi^*(c', \sigma) \subseteq ext(\delta).
         \]
         Therefore,
    $\Phi^*(c,\sigma'') \subseteq ext(\delta)$.\qed
\end{enumerate}

{\lemma \label{lem30b} Let $\delta$ be a p-state and $c$ be a
conditional plan. If ${\mathcal R}^*(c, \delta) = \delta'$ and
$\delta' \neq \bot$, then for every $\sigma \in ext(\delta')$,
$\Phi^*(c, \sigma) \subseteq ext(\delta)$.
}

\proof By induction on $count(c)$, the number of
case plans in $c$.
\begin{enumerate}[$\bullet$]
     \item  Base Case: $count(c)=0$. Then $c$ is a sequence of
     non-sensing actions. The base case follows from Lemma
     \ref{cor1}.
     \item  Inductive Step: Assume that we have proved the
     lemma for $count(c) \leq k$ ($k \geq 0$). We need to prove the lemma for
     $count(c)=k+1$. By the definition of a conditional plan, we
    have two cases:
     \begin{enumerate}[(1)]
         \item $c = a;p$ is a case plan where $a$ is a sensing action and
         $p= case\ (\varphi_1 \rightarrow p_1
         , \ldots ,  \varphi_m \rightarrow p_m)$.
         Since
         $count(c) = 1 + \sum^m_{j=1} count(p_j) \leq
         k+1$, we have that $count(p_i) \leq k$ for $i = 1,\ldots,m$.
        By Definition \ref{d3},
         $$ \bot \neq \delta' = {\mathcal R}^*(c,\delta) =
        {\mathcal R}(a,\{R(\varphi_1 \rightarrow p_1,\delta), \ldots,
                      R(\varphi_m \rightarrow  p_m,\delta)\}).$$
         Let us denote $R(\varphi_i \rightarrow p_i,\delta)$
    by $\delta_i$ ($1 \leq i \leq m$) and $\Delta = \{\delta_1, \ldots,\delta_m\}$.
         We have that $\delta_i \models \varphi_i$ for $1 \leq i \leq m$, and $a$ is applicable in $\Delta$.

         From Theorem \ref{lem2-maintext}, we
         have that
\[
\Phi(a,\sigma) \subseteq \bigcup_{\delta'' \in \Delta} ext(\delta'')
\]
     for every $\sigma \in ext(\delta')$.

    Consider an arbitrary $\sigma' \in \Phi(a,\sigma)$.
    Because of the above relation, we can conclude that there exists
    some $i$, $1 \le i \le m$, such that $\sigma' \in ext(\delta_i)$.
        Because ${\mathcal R}^*(p_i,\delta).T \subseteq R(\varphi_i \rightarrow p_i,\delta).T$ and
         ${\mathcal R}^*(p_i,\delta).F \subseteq R(\varphi_i \rightarrow p_i,\delta).F$,
         ${\sigma}' \in ext(\delta_i)$ implies
         ${\sigma}' \in ext({\mathcal R}^*(p_i,\delta))$.
         Using inductive hypothesis for $count(p_i) \leq
         k$, we have that
    $\Phi^*(p_i,{\sigma}') \subseteq  ext(\delta)$.
    Since this holds for every $\sigma' \in \Phi(a,\sigma)$,
    from Definition \ref{def-extran},
    we conclude that
    $\Phi^*(c,\sigma) \subseteq ext(\delta)$.

         \item $c = a;p$ where $a$ is a  non-sensing action and
    $p$ is a conditional plan. Because $count(c) > 0$, from Definition \ref{defC} we conclude
    that there exists a sequence of non-sensing actions $b_1,\ldots,b_t$
        and a case plan $q$ such that $c = b_1;\ldots;b_t;q$.
        Let $c' = b_1;\ldots;b_t$ and ${\mathcal R}^*(q, \delta) = \delta^*$.
    Using the first case, we can show that for every
    $\sigma'\in ext(\delta^*)$,
    $\Phi^*(q, \sigma') \subseteq ext(\delta)$.
        Furthermore, because
\[
\delta' = {\mathcal R}^*(c, \delta) = {\mathcal R}^*(c', {\mathcal
R}^*(q, \delta))
\]
and Lemma \ref{cor1}, we can show that
    for every
    $\sigma\in ext(\delta)$,
    $\Phi^*(c, \sigma) \subseteq ext(\delta)$.
     \end{enumerate}
\st From cases 1 and 2, the lemma is proved.\qed
\end{enumerate}


We are now ready to prove the soundness of the extended
regression function ${\mathcal R}^*$ with respect to the extended 
progression transition function $\Phi^*$, which is illustrated in the next
figure.

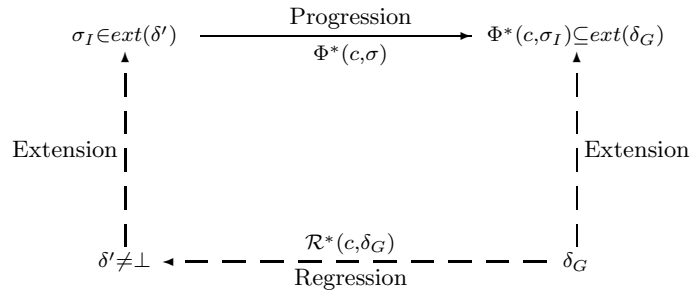
\begin{figure}[htb]
\begin{center}
\setlength{\unitlength}{1cm}
\begin{picture}(6,3.5)
\multiput(5.6,0)(-0.5,0){10}{\line(-1,0){0.3}}
\put(0.5,0){\vector(-1,0){0}}
\put(6,0){\makebox(0,0)[cc]{$_{\delta_G}$}}

\put(3,0.1){\makebox(0,0)[cb]{$_{{\mathcal R}^*(c,\delta_G)}$}}
\put(3,2.9){\makebox(0,0)[ct]{$_{\Phi^*(c,\sigma)}$}}

\multiput(0,0.2)(0,0.5){5}{\line(0,1){0.3}}
\put(0,2.8){\vector(0,1){0}}
\put(0,0){\makebox(0,0)[cc]{$_{\delta' \ne \bot}$}}

\put(0,3){\makebox(0,0)[cc]{$_{\sigma_I \in ext(\delta')}$}}

\put(1,3){\vector(1,0){3.6}}

\multiput(6,0.2)(0,0.5){5}{\line(0,1){0.3}}
\put(6,2.8){\vector(0,1){0}}

\put(6,3){\makebox(0,0)[cc]{$_{\Phi^*(c,\sigma_I)
\subseteq ext(\delta_G)}$}}

\put(3,3.1){\makebox(0,0)[cb]{\footnotesize Progression}}
\put(6.1,1.5){\makebox(0,0)[lc]{\footnotesize Extension}}
\put(-0.1,1.5){\makebox(0,0)[rc]{\footnotesize Extension}}
\put(3,-0.1){\makebox(0,0)[ct]{\footnotesize Regression}}

\end{picture}
\end{center}
     \caption{Soundness of ${\mathcal R}^*$.}
     \label{example-fig-04}
\end{figure}

\begin{theorem} [Soundness of Regression]
\label{lem3-maintext} Let  $P=\langle A,O,I,G \rangle$ be a planning
problem and $c$ be a regression solution of $P$. Then, $c$ is also a
progression solution of $P$, i.e., $\Phi^*(c,\sigma_I) \subseteq
ext(\delta_G)$. 
\end{theorem}

\proof Let $\delta' = {\mathcal R}^*(c,\delta_G)$.
Since $\delta' \neq \bot$ and $\sigma_I \in ext(\delta')$
(Definition \ref{reg-sol}), the conclusion of the theorem follows
immediately from Lemma
\ref{lem30b}.\qed 

\subsection{Completeness Result}

  We now proceed towards a completeness result. Ideally, one
would like to have a completeness result that expresses that for a
given planning problem, any progression solution can also be found
by regression. In our formulation, however, the definition of the
progression function allows an action $a$ to execute in any a-state
$\sigma$ if $a$ is executable in $\sigma$, regardless whether or not
$a$ would add ``new'' information to $\sigma$. In contrast, our
definition of the regression function requires that an action $a$
can only be applied in a p-state (or a set of p-states) if $a$
contributes something to the applied p-state(s) \footnote{Note that
this condition is also applied for regression planning systems such
as \cite{bonet:geffner:hsp-01} and \cite{nguyenetal02}.}. Thus,
given a planning problem $P = \langle A,O,I,G \rangle$, a
progression solution $c$ of $P$ may contain redundant actions or
extra branches. As a result, we may not obtain $c$ via our
regression, i.e. ${\mathcal R}^*(c, \delta_G) = \bot$. To illustrate
this point, let us consider the following two examples.

{\example [Redundancy] \label{EXredundancy-action} Let $P =
\langle \{f,g\}, \{b,c\}, \{f\}, \{g\} \rangle$ be a planning
problem where $c$ is a non-sensing action with $Pre_c = \{f\}$,
$Add_c = \{g\}$, and $Del_c = \emptyset$; $b$ is also a
non-sensing action with $Pre_b = \{g\}$, $Add_b = \{f\}$, and
$Del_b = \emptyset$. Clearly
\[
\begin{array}{l}
p_1 = c \:\:\:\:\:
p_2 = c;b  \:\:\:\:\:
p_3 = c;c
\end{array}
\]
are three progression solutions of $P$. Plan $p_1$ indicates that
$b$ (in $p_2$) and a copy (a.k.a. an instance) of $c$ (in $p_3$) are
redundant.

It is easy to check that $${\mathcal R}^*(p_2, [\{g\}, \emptyset]) = {\mathcal R}^*(p_3, [\{g\}, \emptyset]) = \bot$$ whereas
$${\mathcal R}^*(c, [\{g\}, \emptyset]) = [\{f\},\emptyset].$$
}

{\example [Redundancy] \label{EXredundancy-action2} Let $P =
\langle \{f,g\}, \{b,c\}, \{f\}, \{g\} \rangle$ be a
planning problem. Let $c$ be a sensing action where $Pre_{c} =
\emptyset$, $Sens_{c} = \{f,g\}$; $b$ is a non-sensing action
where $Pre_{b} = \{f, \neg g\}$, $Add_{b} = \{g\}$, and
$Del_{b} = \emptyset$. A plan achieving $g$ is:
\[p = c; case(f \wedge \neg g \rightarrow b, f \wedge g
\rightarrow [\ ], \neg f \wedge \neg g \rightarrow [\ ], \neg f
\wedge g \rightarrow [\ ]).\] Notice that, the
conditions $\neg f \wedge \neg g$ and $\neg f \wedge g$
are always evaluated to false after the execution of $c$ because
$f$ is true before the execution of $c$.
Thus, the two last branches of $p$ are
never used to achieve $g$.

We have that
\[
\begin{array}{l}
{\mathcal R}^*([], [\{g\}, \emptyset]) = [\{g\},\emptyset] \\
{\mathcal R}^*(b, [\{g\}, \emptyset]) = {\mathcal R}(b, [\{g\},\emptyset]) =
[\{f\},\{g\}] \\
\end{array}
\]
We can also verify that
\[
\begin{array}{llllllll}
R(f \wedge \neg g \rightarrow b, [\{g\}, \emptyset]) = [\{f\}, \{g\}] &&
R(f \wedge g \rightarrow [] , [\{g\}, \emptyset]) = [\{f,g\}, \emptyset] \\
R(\neg f \wedge \neg g \rightarrow [], [\{g\}, \emptyset]) = \bot&&
R(\neg f \wedge g \rightarrow [], [\{g\}, \emptyset]) = [\{g\}, \{f\}]
\end{array}
\]
This implies that
\[
{\mathcal R}^*(p, [\{g\}, \emptyset]) =
{\mathcal R}(c, \{[\{f\}, \{g\}], [\{f,g\}, \emptyset], \bot\}) = \bot.
\]
Let $p'$ be the conditional plan obtained from $p$ by removing
the last two branches of $p$, i.e.,

\[p' = c; case(f \wedge \neg g \rightarrow b, f \wedge g
\rightarrow [\ ]).\]

 We can easily check that
${\mathcal R}^*(p', [\{g\}, \emptyset]) = [\{f\},\emptyset] \neq \bot$.
}

\st The above discussion suggests us the following completeness
result: if a conditional plan can be found through progression we
can find an equivalent conditional plan through regression. The plan
found through regression does not have redundancies, both in terms
of extra actions and extra branches. We refer to these notions as
``redundancy'' and ``plan equivalence''. We now formalize these
notions.

\begin{definition}[Subplan] \label{subplan}
Let $c$ be a conditional plan. A conditional plan $c'$ is a {\em
subplan} of $c$ if
\begin{enumerate}[$\bullet$]
     \item  $c'$ can be obtained from $c$ by
\begin{enumerate}[(i)]
\item  removing an instance of a non-sensing action from
     $c$; or
\item removing a case plan or a branch
     $\varphi_i \rightarrow c_i$ from a case plan in $c$; or
\item replacing a case plan
$a;case(\varphi_1 \rightarrow p_1;c_n \ldots,\varphi_m \rightarrow p_m)$
in $c$ with one of its branches $p_i$ for some $i$, $1\le i \le m$; or
\end{enumerate}
     \item  $c'$ is a subplan of $c''$ where $c''$
     is a subplan of $c$.
\end{enumerate}
\end{definition}

The above definition allows us to define redundant plans as follows.

\begin{definition}[Redundancy] \label{redundancy} Let $c$ be a conditional
plan, $\sigma$ be an a-state, and $\delta$ be a p-state. We say
that $c$ {\em contains redundancy} (or is {\em redundant}) with respect to
$(\sigma,\delta)$ if
\begin{enumerate}[(i)]
     \item
         $\Phi^*(c,\sigma) \subseteq ext(\delta)$; and
     \item  there exists a subplan $c'$ of $c$
        with respect to $\sigma$ such that
    $\Phi^*(c',\sigma) \subseteq ext(\delta)$.
\end{enumerate}
\end{definition}

\st Note that, if $c'$ is a subplan of a conditional plan $c$ then
$c' \neq c$. The equivalence of two conditional plans is defined
formally as follows.

\begin{definition} [Equivalent Plan] \label{equivPlan}
Let $\sigma$ be an a-state, $\delta$ be a p-state, and $c$ be a
conditional plan such that
and
$\Phi^*(c,\sigma) \subseteq ext(\delta)$. A conditional
plan $c'$ is equivalent to $c$ with respect to $(\sigma,\delta)$
if $\Phi^*(c',\sigma) \subseteq ext(\delta)$.
\end{definition}

\begin{example} [Equivalence] \label{EXredundancy-action3} Consider the
plans in Example \ref{EXredundancy-action}, we have
that $p_1$ is a subplan of $p_3$ which is equivalent to $p_3$ with
respect to $(\langle \{f\},\emptyset \rangle, [\{g\}, \emptyset])$.

\st Similarly, for planning problem in Example
\ref{EXredundancy-action2}, $p'$ is a subplan of $p$ and is
equivalent to $p$ with respect to $(\langle \{f\},\emptyset \rangle,
[\{g\}, \emptyset])$. 
\end{example}

It is easy to see that if $c'$
and $c''$ are equivalent to $c$ with respect to
$(\sigma,\delta)$ then $c'$ and $c''$ are equivalent with respect
to $(\sigma,\delta)$. To prove the completeness result of our 
regression formulation, we will need to introduce a few
more definitions and notations. Recall that our purpose is to use
regression to find an equivalent conditional plan for a given
progression solution. To do that, we will provide conditions
characterizing when a conditional plan is regressable, i.e. when the
${\mathcal R}^*$ function can be applied on it to produce a p-state.
We refer to conditional plans satisfying such conditions as
\emph{regressable conditional plans}. We will later show that, for a
given progression solution of a planning problem $P$ there exists an
equivalent, regressable conditional plan that is also a regression
solution of $P$.

To define a regressable conditional plan, we begin with some
additional notations. For a non-empty set of fluents  $S =
\{f_1,...,f_k\}$, a binary representation of $S$ is a formula of the
form $l_1 \wedge \ldots \wedge l_k$ where $l_i \in \{f_i, \neg
f_i\}$ for $i= 1, \ldots, k$.

For a non-empty set of fluents $S$, let $BIN(S)$ denote the set of
all different binary representations of $S$. We say a conjunction
$\phi$ of literals is consistent if there exists no fluent $f$ such
that both $f$ and $\neg f$ appear in $\phi$. A set of consistent
conjunctions of literals $\chi = \{\varphi_1, \ldots, \varphi_n\}$
is said to span over some set of fluents $S$ if there exists a
consistent conjunction of literals $\varphi \not \in \chi$, such
that:

\begin{enumerate}[(1)]
     \item $S \cap (\varphi^+ \cup \varphi^-) = \emptyset$ where
     $\varphi^+$ and $\varphi^-$ denote the sets of fluents occurring positive and negative in
     $\varphi$, respectively;
     \item $\varphi_i = \varphi \wedge \psi_i$ where $BIN(S) = \{ \psi_1, \ldots, \psi_n
     \}$.
\end{enumerate}

Notice that for a non-empty set $S$, we can easily check whether
the set $\chi = \{\varphi_1, \ldots, \varphi_n\}$ spans over S. We
say that a set $\chi = \{\varphi_1, \ldots, \varphi_n\}$ is
factorable if it spans over some non-empty set of fluents $S$.

\begin{example} [Getting to Evanston --  Cond't] \label{ex5a}
Consider a set $S= \{\mathit{traffic}$-$bad\}$, a conjunction
$\varphi = on$-$ashland$ and a set of conjunctions $\chi =
\{on$-$ashland \wedge \mathit{traffic}$-$bad, on$-$ashland \wedge
\neg \mathit{traffic}$-$bad\}$.

We have that $BIN(S) = \{\mathit{traffic}$-$bad, \neg
\mathit{traffic}$-$bad\}$
and $\chi$ spans over $S$. 
\end{example}

We can show that for a non-empty set of consistent conjunctions of
literals  $\chi = \{\varphi_1, \ldots, \varphi_n\}$ be a non-empty
set if $\chi$ is factorable, then there exists a unique non-empty
set of fluents $S$ such that $\chi$ spans over $S$. This allows us
to define the notion of regressable plans as follows.

\begin{definition} [Potentially Regressable Case Plan]
\label{Reg-CaseStructure} A case plan $$p = a; case(\varphi_1
\rightarrow c_1, \ldots, \varphi_n \rightarrow c_n)$$ is potentially
regressable if
\begin{enumerate}[(i)]
\item  there exists a non-empty set $\emptyset \ne S_a
\subseteq Sens_a$ such that $\{\varphi_1, \ldots, \varphi_n\}$ spans
over $S_a$, and
\item  for $1 \leq i \leq n$,
     $Sens_a \subseteq (\varphi^+_i \cup \varphi^-_i)$.
\end{enumerate}
\end{definition}

\begin{definition}[Regressable Conditional Plan] \label{RegConditionalPlan}
Let $c$ be a conditional plan, $\sigma$ be an a-state, and $\delta$
be a p-state. We say $c$ is regressable with respect to
$(\sigma,\delta)$ if
\begin{enumerate}[(i)]
\item  every case plan occurring in $c$ is
potentially regressable,
\item  $\Phi^*(c,\sigma) \subseteq ext(\delta)$, and
\item  $c$ is not redundant with respect to
$(\sigma,\delta)$.
\end{enumerate}
\end{definition}

We will now prove a series of lemmae that will be used in the proof of 
the completeness of ${\mathcal R}^*$. Lemma \ref{factorable} is about the uniqueness of 
a set of literals over which a factorable set of conjunctions spans.
Lemmae \ref{lem311}-\ref{lem31} state that the regressable property 
of a sequence of non-sensing actions 
is maintained by the function ${\mathcal R}^*$.
Lemma \ref{lem31ab}-\ref{lem31b} extend this result to regressable 
conditional plans. Lemmae \ref{lemAdd2}-\ref{lemAdd3} show that for each 
progression solution there exists an equivalent regressable plan which 
can be found through regression.

{\lemma \label{factorable} Let $\chi = \{\varphi_1, \ldots,
\varphi_n\}$ be a non-empty set of consistent conjunctions of
literals. If $\chi$ is factorable, then there exists a unique
non-empty set of fluents $S$ such that $\chi$ spans over $S$.
}

\proof Since $\chi$ is factorable, there exists a
non-empty set of fluents $S$ such that $\chi$ spans over $S$, i.e.
there exists $\varphi$ such that $\varphi_i = \varphi \wedge \psi_i$
where $\psi_i \in BIN(S)$ for $i=1, \ldots,n$ and $BIN(S) =
\{\psi_1,\ldots,\psi_n\}$. Assume that $S$ is not unique. This means
that there exists a non-empty set $S' \neq S$ such that $\chi$ spans
over $S'$, i.e. there exists $\varphi'$ such that $\varphi_i =
\varphi' \wedge \psi'_i$ where $\psi'_i \in BIN(S')$ for $i=1,
\ldots,n$.

\st Consider $f \in S \setminus S'$. For every $1 \leq i \leq n$, we
have that $\varphi_i = \varphi' \wedge \psi'_i$. Since $f \not \in
S'$ and $\varphi_i$ is consistent ($1 \leq i \leq n$), $f$ must
occur either positively or negatively in $\varphi'$. This means that
$f$ occurs either positively or negatively in all $\varphi_i$ for $1
\leq i \leq n$.

\st Consider the case that $f$ occurs positively in all $\varphi_i$
for $1 \leq i \leq n$ [*]. Since $f \in S$, there exists a binary
representation $\psi_j \in BIN(S)$ ($1 \leq j \leq n$) such that $f$
appears negatively in $\psi_j$ i.e. $f$ appears negatively in
$\varphi_j$. This contradicts with [*]. Similarly we can show a
contradiction in the case that $f$ occurs negatively in all
$\varphi_i$ for $1 \leq i \leq n$. We conclude that $S$ is
unique.\qed 

{\lemma \label{lem311} Let $\sigma$ be an a-state, $\delta$ be a
p-state, and $c=a_1; \ldots;a_n$ ($n \geq 1$) be a sequence of
non-sensing actions. Assume that $c$ is regressable with respect to
$(\sigma,\delta)$. Then, ${\mathcal R}^*(a_n,\delta) = \delta'$,
$\delta' \neq \bot$, and $c' = a_1;\ldots;a_{n-1}$ is regressable
with respect to $(\sigma,\delta')$.  
}

\proof By induction on $n$.

\begin{enumerate}[$\bullet$]
     \item  Base Case: $n=1$. Similar to the inductive step,
we can show that $a_1$ is applicable in $\delta$. Let
$\delta'={\mathcal R}(a_1,\delta)$ and $\Phi(a_1,\sigma) =
\{\sigma'\}$. We have that, $\delta'.T = (\delta.T \setminus
Add_{a_1}) \cup Pre_{a_1}^+$ and $\sigma'.T = (\sigma.T \setminus
Del_{a_1}) \cup Add_{a_1}$. Using the facts $\sigma' \in
ext(\delta)$, $Add_{a_1} \cap Del_{a_1} = \emptyset$, and the above
equations, we can show that $\delta'.T \subseteq \sigma.T$.
Similarly, $\delta'.F \subseteq \sigma.F$. Since $[\ ]$ is not
redundant with respect to $(\sigma,\delta')$, we have that $[\ ]$ is
a plan that is regressable with respect to $(\sigma,\delta')$.

     \item  Inductive Step: Assume that we have proved the lemma
     for $0 < n \le k$. We need to prove the lemma for $n=k+1$.

     Let $\Phi^*(a_1; \ldots; a_k,\sigma) = \{\sigma_k\}$, we have
     that
     \[\Phi^*(c,\sigma) = \Phi(a_{k+1},\sigma_k) = \{\sigma'\} \subseteq
     ext(\delta).\]

     We will prove that (1) $a_{k+1}$ is applicable in $\delta$,
     (2) ${\mathcal R}(a_{k+1},\delta) = \delta^* \ne \bot$ and
      $\sigma_k \in ext(\delta^*)$, and (3) $c'=a_1; \ldots; a_k$ is
     regressable with respect to $(\sigma,\delta^*)$.

\medskip
     \begin{enumerate}[--]
         \item Proof of (1):
     We first show that
     $Add_{a_{k+1}} \cap \delta.T \neq \emptyset$ or
     $Del_{a_{k+1}} \cap \delta.F \neq \emptyset$.
     Assume the contrary, $Add_{a_{k+1}} \cap \delta.T = \emptyset$ and
     $Del_{a_{k+1}} \cap \delta.F = \emptyset$.
     By Definition \ref{def-tran}, we have that
     $$\sigma'.T = (\sigma_k.T \setminus Del_{a_{k+1}}) \cup Add_{a_{k+1}}$$
      and
     $$\sigma'.F = (\sigma_k.F \setminus Add_{a_{k+1}}) \cup Del_{a_{k+1}}.$$
     Since $\sigma' \in ext(\delta)$, we have $\delta.T \subseteq
     \sigma'.T$. By our assumption, $Add_{a_{k+1}} \cap \delta.T = \emptyset$,
     we must have that $\delta.T = \delta.T \setminus Add_{a_{k+1}} \subseteq \sigma'.T \setminus
     Add_{a_{k+1}}$. Because for arbitrary sets $X,Y$,
     $(X \cup Y) \setminus Y = X \setminus (X \cap Y)$, we have that
     \[\sigma'.T \setminus Add_{a_{k+1}} =
              ((\sigma_k.T \setminus Del_{a_{k+1}}) \cup Add_{a_{k+1}})
     \setminus Add_{a_{k+1}} =
     \]
     \[(\sigma_k.T \setminus Del_{a_{k+1}}) \setminus ((\sigma_k.T \setminus Del_{a_{k+1}}) \cap Add_{a_{k+1}}) \subseteq \sigma_k.T ,\]
     i.e. $ \delta.T \subseteq \sigma_k.T \setminus Del_{a_{k+1}}$.
     This shows that $\delta.T \subseteq \sigma_k.T$. Similarly,
    we can show that $\delta.F \subseteq
     \sigma_k.F$. We conclude that $\sigma_k \in ext(\delta)$, i.e.
     $c$ is redundant with respect to $(\sigma,\delta)$.
     This is a contradiction. Therefore, $Add_{a_{k+1}} \cap \delta.T \neq \emptyset$ or
     $Del_{a_{k+1}} \cap \delta.F \neq \emptyset$.
\hfill(i)

     Since $\sigma' \in
     ext(\delta)$, we have $\delta.T \subseteq \sigma'.T$ and $\delta.F
     \subseteq \sigma'.F$. As $a_{k+1}$ is executable in $\sigma_k$, we have
     $Add_{a_{k+1}} \cap \sigma'.F = \emptyset$ and $Del_{a_{k+1}} \cap
     \sigma'.T = \emptyset$. This concludes that $Add_{a_{k+1}} \cap \delta.F =
     \emptyset$ and $Del_{a_{k+1}} \cap \delta.T = \emptyset$.
\hfill(ii)

     Now, assume that there exists $f \in Pre^+_{a_{k+1}} \cap
     \delta.F$ and $f \not \in Del_{a_{k+1}}$. By Definition
     \ref{def-tran}, it's easy to see
     that $f \in \sigma'.T$ and $f \in \sigma'.F$. This is a
     contradiction, therefore $Pre^+_{a_{k+1}} \cap \delta.F \subseteq
     Del_{a_{k+1}}$. Similarly, we can show that $Pre^-_{a_{k+1}} \cap
     \delta.T \subseteq
     Add_{a_{k+1}}$.
\hfill(iii).

   From (i), (ii), and (iii) we conclude
     that $a_{k+1}$ is applicable in $\delta$.

\medskip
     \item Proof of (2):
     Because $a_{k+1}$ is applicable in $\delta$, we have that
     ${\mathcal R}(a_{k+1},\delta) = \delta^*$ for some
    partial state $\delta^* \neq \bot$.
     We will show that $\sigma_k \in ext(\delta^*)$.
     From the fact $\sigma' \in ext(\delta)$, by Definition \ref{def-tran},
     we have
     \[\delta.T \subseteq \sigma'.T = (\sigma_k.T \setminus Del_{a_{k+1}})
     \cup
     Add_{a_{k+1}}\] and
     \[\delta.F \subseteq \sigma'.F = (\sigma_k.F \setminus Add_{a_{k+1}})
     \cup
     Del_{a_{k+1}}.\]

     By Definition \ref{reg-nonsensing} we have
     \[\delta^*.T = (\delta.T \setminus Add_{a_{k+1}}) \cup
     Pre^+_{a_{k+1}}\] and
     \[\delta^*.F = (\delta.F \setminus Del_{a_{k+1}}) \cup
     Pre^-_{a_{k+1}}.\]
     Since $a_{k+1}$ is executable in $\sigma_k$, we have that
     $Pre^+_{a_{k+1}} \subseteq \sigma_k.T$
     and $Pre^-_{a_{k+1}} \subseteq \sigma_k.F$. Therefore, to prove
     that $\delta^*.T = (\delta.T \setminus Add_{a_{k+1}}) \cup
     Pre^+_{a_{k+1}} \subseteq \sigma_k.T$, we only need to show that
     $\delta.T \setminus Add_{a_{k+1}} \subseteq \sigma_k.T$. As
     $\delta.T \subseteq (\sigma_k.T \setminus Del_{a_{k+1}}) \cup
     Add_{a_{k+1}}$, we have
     \[\delta.T \setminus Add_{a_{k+1}} \subseteq ((\sigma_k.T \setminus
     Del_{a_{k+1}}) \cup Add_{a_{k+1}}) \setminus Add_{a_{k+1}}.\]
     From the proof of item (1), we have that $((\sigma_k.T \setminus
     Del_{a_{k+1}}) \cup Add_{a_{k+1}}) \setminus Add_{a_{k+1}} \subseteq
     \sigma_k.T$. This concludes that $\delta.T \setminus Add_{a_{k+1}} \subseteq
     \sigma_k.T$. Similarly, we can show that $\delta.F \setminus
     Del_{a_{k+1}} \subseteq
     \sigma_k.F$, i.e., $\sigma_k \in ext(\delta^*)$ or $\{\sigma_k\}
     \subseteq ext(\delta^*)$.

\medskip
     \item Proof of (3):
     Suppose that $c'$ is redundant with respect to $(\sigma,\delta^*)$.
     By Definition \ref{redundancy}, there exists a subplan $c''$ of $c$
     such that
     $\Phi^*(c'', \sigma) = \{\sigma''\} \subseteq ext(\delta^*)$.
     By Theorem \ref{lem1-maintext}, we have that $\Phi(a_{k+1},\sigma'') \subseteq
     ext(\delta)$. Since \[\Phi^*(c''; a_{k+1}, \sigma) =
     \Phi(a_{k+1}, \sigma'') \subseteq ext(\delta),\]
     we have that $c$ is redundant with respect to
     $(\sigma,\delta)$. This
     contradicts with the assumption that $c$ is not redundant
     with respect to $(\sigma,\delta)$. Since $c'$ has no case
     plan, this concludes that $c'$ is not
     redundant with respect to $(\sigma,\delta^*)$. Since
     $\Phi^*(a_1; \ldots; a_k,\sigma) = \{\sigma_k\} \subseteq ext(\delta^*)$
     we have that $c'$ is regressable with respect to $(\sigma,\delta^*)$.\qed
\end{enumerate}
\end{enumerate}

{\lemma \label{lem31} Let $\sigma$ be an a-state and $\delta$ be a
p-state. Let $c=a_1; \ldots;a_n$ be a sequence of non-sensing
actions that is regressable with respect to $(\sigma,\delta)$. Then,
there exists some p-state $\delta^* \neq \bot$ such that  ${\mathcal
R}^*(c,\delta) = \delta^*$ and $\sigma \in ext(\delta^*)$.
}

\proof By induction on $n$.
\begin{enumerate}[$\bullet$]
     \item  Base Case: $n=0$. Then $c$ is an empty sequence of
     non-sensing actions. The base case follows from Definition \ref{d3}
     (with $\delta^* = \delta$ and $[\ ]$ is not redundant with respect
    to $(\sigma,\delta)$).

     \item  Inductive Step: Assume that we have proved the
     lemma for $0 \le n \le k$. We need to prove the lemma for
     $n=k+1$.
     It follows from Lemma \ref{lem311}
     that $\delta' = {\mathcal R}(a_{k+1},\delta)$,
     $\delta' \ne \bot$, and
     $c' = a_1;\ldots;a_{k}$ is a plan that is regressable
     with respect to $(\sigma,\delta')$.
     By inductive hypothesis,  we have that
     ${\mathcal R}^*(c',\delta') =\delta^* \ne \bot$
     and $\sigma \in ext(\delta^*)$.
     The inductive step follows from this and the fact
     ${\mathcal R}^*(c,\delta) = {\mathcal R}^*(c',{\mathcal R}(a_{k+1},\delta))$.\qed
\end{enumerate}

\lemma \label{lem31ab} Let $\sigma$ be an a-state, $a$ be a sensing
action which is executable in $\sigma$. Let $S_a = Sens_a \setminus
\sigma$. Then, we have that
\begin{enumerate}[(1)]
\item $\Phi(a,\sigma) = \{\sigma_1,\ldots, \sigma_m\}$ where $m=2^{|S_a|}$,

\item $a$ is strongly
applicable in $\Delta = \{\delta_1, \ldots, \delta_m\}$ where
$\delta_i = [\sigma_i.T, \sigma_i.F]$, $i=1,\ldots,m$, and 

\item ${\mathcal R}(a, \Delta) = [\sigma.T, \sigma.F]$.
\end{enumerate}

\proof\hfill
\begin{enumerate}[(1)]
\item From Definition \ref{def-tran}, we have that
\[\bot \not \in \Phi(a,\sigma) = \{\sigma' | Sens_a \setminus \sigma = \sigma' \setminus \sigma\}\]
and, for every $\sigma' \in \Phi(a,\sigma)$, $\sigma' \setminus
\sigma = (\sigma'.T \setminus \sigma.T) \cup (\sigma'.F \setminus
\sigma.F)$. Denote $\sigma'.T \setminus \sigma.T$ by $P$ and
$\sigma'.F \setminus \sigma.F$ by $Q$, we have that $(P,Q)$ is a
partition of $S_a$. Since there are $2^{|S_a|}$ partitions of $S_a$,
we have that $m \leq 2^{|S_a|}$. Furthermore, for a partition
$(P,Q)$ of $S_a$ there exists an a-state $\sigma' = \langle P \cup
\sigma.T, Q \cup \sigma.F \rangle \in \Phi(a,\sigma)$ because
$\sigma' \setminus \sigma = P \cup Q$. Therefore $2^{|S_a|} \leq m$.
We conclude that $m = 2^{|S_a|}$.

\item We first show that $\Delta$ is proper with respect
to $S_a$, i.e. $S_a$ is a sensed set of $\Delta$ with respect to
$a$. Indeed, by Definition \ref{def-tran} and the proof of (1), we
have that the first three conditions of Definition \ref{sensed-Set}
are satisfied. The fourth condition of Definition \ref{sensed-Set}
is satisfied because we have that $\delta_i.T \setminus S_a =
\sigma_i.T \setminus S_a = \sigma.T$ and $\delta_i.F \setminus S_a =
\sigma_i.F \setminus S_a = \sigma.F$ ($1 \leq i \leq m$). Therefore,
we conclude that $p(a,\Delta) = S_a$.

Since $a$ is an action that is executable in $\sigma$ we have
that $(Pre^+_a \cup Pre^-_a) \cap Sens_a = \emptyset$ and $Pre^+_a
\cap \sigma.F = \emptyset$, $Pre^-_a \cap \sigma.T = \emptyset$,
therefore $Pre^+_a \cap \delta_i.F = \emptyset$, $Pre^-_a \cap
\delta_i.T = \emptyset$ ($1 \leq i \leq m$). By Definition
\ref{app-sensing-strong}, we conclude that $a$ is strongly
applicable in $\Delta$.

\item Since $a$ is executable in $\sigma$, we have that
$Pre^+_a \subseteq \sigma.T$ and $Pre^-_a \subseteq \sigma.F$. From
the proof of (2), $\delta_i.T \setminus S_a = \sigma.T$ and
$\delta_i.F \setminus S_a = \sigma.F$ ($1 \leq i \leq m$). The proof
follows from Definition \ref{reg-sensing}.\qed
\end{enumerate}

{\lemma \label{lem31b-son} Let $\sigma$ be an a-state, $\delta$ be a
p-state, and $c = \alpha; c'$ is a conditional plan where $\alpha$
is a non-empty sequence of non-sensing actions and $c' = a;
case(\varphi_1 \rightarrow p_1,\ldots,
           \varphi_m \rightarrow p_m)$.
If $c$ is regressable with respect to $(\sigma,\delta)$, then

\begin{enumerate}[(1)]
\item  there exists some a-state $\sigma_1 \ne \bot$ such that
$\Phi^*(\alpha, \sigma) = \{\sigma_1\}$;

\item  $m = 2^{|S_a|}$ where $S_a = Sens_a \setminus
\sigma_1$;

\item  $\{\varphi_1,\ldots,\varphi_m\}$ spans over
$S_a$;

\item  For each $i$, $1 \le i \le m$, there exists a
unique a-state $\sigma' \in \Phi(a, \sigma_1)$ such that $p_i$ is
regressable with respect to $(\sigma',\delta)$. 
\end{enumerate}
}

\proof\hfill

\begin{enumerate}[(1)]
\item  By Definition \ref{def-extran}, we have that
\[
\Phi^*(c, \sigma) = \bigcup_{\sigma' \in  \Phi^*(\alpha, \sigma)}
\Phi^*(c', \sigma').
\]
Since $c$ is regressable with respect to $(\sigma,\delta)$ we have
that $\bot\not\in \Phi^*(c, \sigma)$. This implies that $\bot
\not\in \Phi^*(\alpha, \sigma)$. Furthermore, because $\alpha$ is a
sequence of non-sensing actions, we conclude that there exists  some
a-state $\sigma_1 \ne \bot$. such that $\Phi^*(\alpha, \sigma) =
\{\sigma_1\}$.

\item  By definition of $S_a$ we conclude that $S_a$ is
the set of fluents that belong to $Sens_a$ which are unknown in
$\sigma_1$. By Definition \ref{def-tran}, we conclude that $\Phi(a,
\sigma_1)$ consists of $2^{|S_a|}$ elements where for each $\sigma'
\in \Phi(a, \sigma_1)$, $\sigma' \setminus \sigma_1 = S_a$. Because
\[
\bot \not\in \Phi^*(c, \sigma) = \bigcup_{\sigma' \in  \Phi(a,
\sigma_1)} E(case(\varphi_1 \rightarrow p_1,\ldots,
           \varphi_m \rightarrow p_m), \sigma')
\]
we conclude that for each $\sigma' \in \Phi(a, \sigma_1)$ there
exists one $j$, $1 \le j \le m$, such that $\varphi_j$ is satisfied
in $\sigma'$. Since $\varphi$'s are mutual exclusive we conclude
that for each $j$, $1 \le j \le m$, there exists at most one
$\sigma' \in \Phi(a, \sigma_1)$ such that $\varphi_j$ is satisfied
in $\sigma'$. This implies that $m \ge 2^{|S_a|}$. The
non-redundancy property of $c$ implies that $m \ge 2^{|S_a|}$. Thus,
$m = 2^{|S_a|}$.

\item  Since $c$ is regressable with respect to
$(\sigma,\delta)$ we have that $a; (case(\varphi_1 \rightarrow
p_1,\ldots, \varphi_m \rightarrow p_m)$ is potentially regressable.
This implies that $\{\varphi_1,\ldots,\varphi_m\}$ spans over a set
of fluents $S \subseteq Sens_a$ and there exists a $\varphi$ such
that for every $i$, $\varphi_i = \psi_i \wedge \varphi$ where
$\psi_i \in BIN(S)$ and $S \cap (\varphi^+ \cup \varphi^-) =
\emptyset$. From Lemma \ref{factorable} we know that $S$ is unique.
We will show now that $S = S_a$. Assume the contrary, $S \ne S_a$.
We consider two cases:

\begin{enumerate}[$\bullet$]
\item $S \setminus S_a \ne \emptyset$. Consider a fluent $f \in S
\setminus S_a$. Because $\{\varphi_1,\ldots,\varphi_m\}$ spans over
$S$, there exists some $i$ such that $f$ occurs positively in
$\varphi_i$. From the proof of the previous item and the fact that
$f \not\in S_a$, we conclude that $f$ must be true in $\sigma_1$
(otherwise, we have that the subplan $c'$ of $c$, obtained by
removing the branch $\varphi_i \rightarrow p_i$, satisfies $\bot
\not\in \Phi^*(c',\sigma) \subseteq ext(\delta)$, which implies that
$c$ is redundant with respect to $(\sigma,\delta)$). Similarly,
there exists some $j$ such that $f$ occurs negatively in
$\varphi_j$, and hence, $f$ must be false in $\sigma_1$. This is a
contradiction. Thus, this case cannot happen.

\item $S_a \setminus S \ne \emptyset$. Consider a fluent $f \in
S_a \setminus S$. Again, from the fact that $c$ is regressable with
respect to $(\sigma,\delta)$, we conclude that $f$ occurs either
positively or negatively in $\varphi_i$. Because $f \not\in S$, we
have that $f$ occurs in $\varphi$, and hence, $f$ occurs positively
or negatively in all $\varphi_i$. In other words, $f$ is true or
false in every $\sigma' \in \Phi(a,\sigma_1)$. Thus, $f$ is true or
false in $\sigma_1$. This contradicts the fact that $f \in S_a =
Sens_a \setminus \sigma_1$. Thus, this case cannot happen too.
\end{enumerate}

\noindent The above two cases imply that $S_a = S$. This means that
$\{\varphi_1,\ldots,\varphi_m\}$ spans over $S_a$.

\item  Consider an arbitrary $i$, $1 \le i \le m$. From
the proof of the second item, we know that there exists a unique
$\sigma' \in \Phi(a,\sigma_1)$ such that $\varphi_i$ is satisfied by
$\sigma'$. We will show now that $p_i$ is regressable with respect
to $(\sigma',\delta)$. From the fact that $c$ is regressable, we
conclude that every case plan in $p_i$ is potentially regressable.
Furthermore, because $\Phi^*(p_i, \sigma') \subseteq \Phi^*(c,
\sigma)$, we have that $\bot \not\in \Phi^*(p_i, \sigma') \subseteq
ext(\delta)$. Thus, to complete the proof, we need to show that
$p_i$ is not redundant with respect to $(\sigma', \delta)$. Assume
the contrary, there exists a subplan $p'$ of $p_i$ such that $\bot
\not\in \Phi^*(p', \sigma') \subseteq ext(\delta)$. This implies
that the subplan $c'$ of $c$, obtained by replacing $p_i$ with $p'$,
will satisfy that $\bot \not\in \Phi^*(c', \sigma) \subseteq
ext(\delta)$, i.e., $c$ is redundant with respect to $(\sigma,
\delta)$. This contradicts the condition of the lemma, i.e., our
assumption is incorrect. Thus, $p_i$ is not redundant with respect
to $(\sigma', \delta)$, and hence, $p_i$ is regressable with respect
to $(\sigma', \delta)$.\qed
\end{enumerate}

{ \lemma \label{lem31b} Let $\sigma$ be an a-state, $\delta$ be a
p-state, and $c$ is a conditional plan that is regressable with
respect to $(\sigma,\delta)$. Then, there exists some p-state
$\delta' \neq \bot$ such that ${\mathcal R}^*(c,\delta) = \delta'$
and $\sigma \in ext(\delta')$. 
}

\proof By induction on $count(c)$, the number of
case plans in $c$.
\begin{enumerate}[$\bullet$]

\item  Base Case: $count(c)=0$. Then $c$ is a sequence
of non-sensing actions. The base case follows from Lemma
\ref{lem31}.

\item  Inductive Step:
     Assume that we have proved the lemma
     for $count(c) \leq k$. We need to prove the lemma for
     $count(c)=k+1$.
Since $c$ is a conditional plan, we have that $c = \alpha; c'$ where
$\alpha$ is a sequence of non-sensing actions and $c' = a; p$ and
$p= case\ (\varphi_1 \rightarrow p_1
         , \ldots ,  \varphi_m \rightarrow p_m)$.
Because $\alpha$ is a sequence of non-sensing actions we have that
$\Phi^*(\alpha, \sigma)$ is a singleton. Let $\Phi^*(\alpha, \sigma)
= \{\sigma_1\}$.

\st Let $S_a = Sens_a \setminus \sigma_1$. Since $c$ is not
redundant with respect to $(\sigma,\delta)$ we conclude that $S_a
\ne \emptyset$.

\st It follows from the fact that  $c$ is regressable with respect
to $(\sigma,\delta)$ and Lemma \ref{lem31b-son} that $\{\varphi_1,
\ldots, \varphi_m\}$ spans over $S_a$ and for every $i$, $1 \le i
\le m$, there exists a unique $\sigma' \in \Phi(a, \sigma_1)$ such
that $p_i$ is regressable with respect to $(\sigma',\delta)$. By
inductive hypothesis for $p_i$, we conclude that ${\mathcal
R}^*(p_i, \delta) = \delta_i \ne \bot$ and $\sigma' \in
ext(\delta_i)$. Because $\varphi_i$ is satisfied by $\sigma'$ we
have that $R(\varphi_i \rightarrow p_i,\delta) = [\delta_i.T \cup
\varphi_i^+, \delta_i.F \cup \varphi_i^-]$ is consistent and hence
$R(\varphi_i \rightarrow p_i,\delta) \ne \bot$. This also implies
that $\sigma' \in ext(R(\varphi_i \rightarrow p_i,\delta))$ and
$R(\varphi_i \rightarrow p_i,\delta) \ne R(\varphi_j \rightarrow
p_j,\delta)$ for $i \ne j$.

\st Let $\Delta = \{R(\varphi_i \rightarrow p_i ,\delta) \mid
i=1,\ldots,m\}$. We will show next that $a$ is applicable in
$\Delta$. Consider $\Delta' = \Phi(a, \sigma_1)$, we have that for
each $i$, $1 \le i \le m$, there exists one $\sigma' \in \Delta'$
and $\sigma' \in ext(R(\varphi_i \rightarrow p_i,\delta))$. It
follows from Lemma \ref{lem31ab} that $a$ is strongly applicable in
$\Delta'$. Thus, $a$ is applicable in $\Delta$.

\st By definition of ${\mathcal R}$, we have that

\[
\begin{array}{ll}
{\mathcal R}(a,\Delta)  =
[((\bigcup_{i=1}^m R(\varphi_i \rightarrow p_i,\delta).T) \setminus S_a) \cup Pre^+_a , \\
\hspace*{1in}  ((\bigcup_{i=1}^m R(\varphi_i \rightarrow
p_i,\delta).F) \setminus S_a) \cup Pre^-_a
        ] = \delta^* \ne \bot.
\end{array}
\]

Since $a$ is executable in $\sigma_1$, from Lemma \ref{lem31ab}, and
the fact that for each $\sigma' \in \Phi(a,\sigma_1)$ there exists
an $i$ such that $\sigma' \in ext(R(\varphi_i \rightarrow
p_i,\delta))$, we can conclude $\sigma_1 \in ext(\delta^*)$.

\st To continue our proof, we will now show that $q = \alpha$ is not
redundant with respect to $(\sigma, \delta^*)$. Assume the contrary,
there exists a subplan $q'$ of $q$ such that $\Phi^*(q', \sigma)
\subseteq ext(\delta^*)$. This, together with the fact that
${\mathcal R}^*(c', \delta) = \delta^*$ and Theorem \ref{lem3-maintext}
implies that $\Phi^*(c'', \sigma) \subseteq ext(\delta)$ for $c'' =
q';c'$, i.e., $c$ is redundant with respect to $(\sigma,\delta)$.
This contradicts the assumption of the lemma, i.e., we have proved
that $q$ is not redundant with respect to $(\sigma, \delta^*)$.

\st Applying the inductive hypothesis for the plan $q$ and
$(\sigma,\delta^*)$, we have that ${\mathcal R}^*(q, \delta^*) =
\delta' \ne \bot$ and $\sigma \in ext(\delta')$. The inductive
hypothesis is proved because ${\mathcal R}^*(c, \delta) = {\mathcal
R}^*(q, \delta^*)$.\qed
\end{enumerate}

\st {\lemma \label{lemAdd2} Let $\sigma$ be an a-state, $\delta$ be
a p-state, and $c$ be a sequence of non-sensing actions such that
$\Phi^*(c,\sigma) \subseteq ext(\delta)$. Then, there exists a
subplan $c'$ of $c$ that is not redundant with respect to
$(\sigma,\delta)$ and $c'$ is equivalent to $c$ with respect to
$(\sigma,\delta)$.
}

\proof Notice that the length of $c$ is
finite\footnote{
   By this we mean that $c$ is given and hence its length (the number of
   actions in $c$) is finite.
}. Consider two cases:
\begin{enumerate}[$\bullet$]
     \item  Case (i): $c$ is not redundant with respect to $(\sigma,\delta)$.

     It's easy to see that $c'=c$ satisfies the condition of the lemma.

     \item  Case (ii): $c$ is redundant with respect to $(\sigma,\delta)$.

     By definition of redundancy, there exists a subplan of $c$ which
     are equivalent to $c$ with respect to $(\sigma,\delta)$.
     Let $c'$ be a subplan of $c$ which is equivalent to $c$
     with respect to $(\sigma,\delta)$ whose length is minimal among all subplans
     which is equivalent to $c$ with respect to $(\sigma,\delta)$.  To prove the lemma,
     it suffices to show that $c'$ is not redundant with respect to $(\sigma,\delta)$.
     Assume the contrary, there exists a subplan $c''$ of $c'$ which
     is equivalent to $c$ with respect to $(\sigma,\delta)$. Trivially,
     the number of actions in $c''$ is smaller than the number
     of actions in $c'$. By definition, we have that $c''$ is also a subplan of $c$
     which is equivalent
     to $c$ with respect to $(\sigma,\delta)$. This contradicts the fact that
     $c'$ has the minimal length among all subplans of $c$ which are equivalent to $c$.
      So, we conclude that
      $c'$ is not redundant with respect to $(\sigma,\delta)$. The lemma is
      proved.\qed
\end{enumerate}

\st {\lemma \label{lemAdd3-son} Let $\sigma$ be an a-state and $c =
a; case\ (\varphi_1 \rightarrow p_1
         , \ldots ,  \varphi_m \rightarrow p_m)$
be a case plan such that $\bot \not \in \Phi^*(c,\sigma)$. Then, if
$Sens_a \setminus \sigma \ne \emptyset$, there exists a potentially
regressable plan $c' = a; case\ (\varphi_1' \rightarrow p'_1
         , \ldots ,  \varphi_n' \rightarrow p'_n)$
such that $\Phi^*(c,\sigma) = \Phi^*(c', \sigma)$. 
}

\proof We prove the lemma by constructing $c'$. Let
$S = \{\varphi_1,\ldots,\varphi_m\}$ and $S_a = Sens_a \setminus
\sigma$. Let $L = \{f \mid f \in S_a\} \cup \{\neg f \mid f \in
S_a\}$. First, observe that because of $\bot \not \in
\Phi^*(c,\sigma)$ we have that $a$ is executable in $\sigma$.
Furthermore, for each $\sigma' \in \Phi(a,\sigma)$ there exists one
$\varphi_i \in S$ such that $\varphi_i$ is satisfied in $\sigma'$.
Without loss of generality, we can assume that for each $\varphi_i
\in S$, there exists (at least) one $\sigma' \in \Phi(a,\sigma)$
such that $\varphi_i$ is satisfied in $\sigma'$.

\st It is easy to see that for each $i$, we can write $\varphi_i =
\psi_i \wedge \chi_i$ where $\psi_i$ is the conjunction of literals
occurring in $\varphi_i$ and belonging to $L$ and $\chi_i$ is the
conjunction of literals that do not belong to $L$. From the above
observation, we have that $\chi_i$ is satisfied by $\sigma$. So,
$\varphi = \wedge_{i=1}^m \chi_i$ holds in $\sigma$. Thus, the
conditional plan $c_1 = a; case\ (\varphi'_1 \rightarrow p_1 ,
\ldots ,  \varphi'_m \rightarrow p_m)$ where $\varphi'_i = \psi_i
\wedge \varphi$ satisfies that
  $\Phi^*(c,\sigma) = \Phi^*(c_1, \sigma)$.

\st Since $\psi_i$ is a consistent conjunction of literals from $L$
and $\psi_i$'s are mutual exclusive, there exists a partition
$(S_1,\ldots,S_m)$ of $BIN(S_a)$ such that for every $\eta \in S_i$,
$\eta = \psi_i \wedge \eta'$. Let

\[
\begin{array}{rll}
c_2 = a; case( &  \\
 & & \gamma^1_1 \rightarrow p_1 , \ldots ,  \gamma^{|S_1|}_1 \rightarrow p_1, \\
 & & \gamma^1_2 \rightarrow p_1 , \ldots ,  \gamma^{|S_2|}_2 \rightarrow p_2, \\
 & & \ldots  \\
 & & \gamma^1_m \rightarrow p_1 , \ldots ,  \gamma^{|S_m|}_m \rightarrow p_m, \\
) & & \\
\end{array}
\]
where 
\[\eqalign{
  \gamma^j_i
&=\eta^j_i \wedge \varphi \wedge \gamma\cr
  S_i
&=\{\eta^1_i,\ldots,\eta^{|S_i|}_i\}\quad\hbox{for $i=1,\ldots,m$,
    and}\cr
  \gamma
&=\bigwedge{}_{f \in Sens_a \cap \sigma.T}\, f \wedge
         \bigwedge{}_{f \in Sens_a \cap \sigma.F}\, \neg f\ .
}
\]
We have that $\Phi^*(c,\sigma) = \Phi^*(c_2, \sigma)$. It is easy to
see that the set $\{\gamma^1_1,\ldots,\gamma^{|S_m|}_m\}$ spans over
$S_a$ and $Sens_a \subseteq (\gamma^j_i)^+ \cup (\gamma^j_i)^-$.
Thus, $c_2$ is potentially regressable. The lemma is proved with $c'
= c_2$.\qed

\st {\lemma \label{lemAdd3} Let $\sigma$ be an a-state, let $\delta$
be a p-state, and let $c$ be a conditional plan such that
$\Phi^*(c,\sigma) \subseteq ext(\delta)$. There exists a plan $c'$
such that $c'$ is regressable with respect to $(\sigma,\delta)$ and
$c'$ is equivalent to $c$ with respect to $(\sigma,\delta)$. 
}

\proof
By induction on $count(c)$, the number of case plans in $c$.

\begin{enumerate}[$\bullet$]
\item  Base case: $count(c) = 0$

     This follows from Lemma \ref{lemAdd2}.
\item  Inductive Step: Assume that we have proved the
     lemma for $count(c) \leq k$. We need to prove the lemma for
     $count(c)=k+1$.

     By construction of $c$, we have two cases
     \begin{enumerate}[(1)]
     \item $c= a; p$ where
     $p= case\ (\varphi_1 \rightarrow p_1 , \ldots ,  \varphi_m \rightarrow p_m)$.
     Here, we have two cases.
     \begin{enumerate}[(a)]
     \item $Sens_a \setminus \sigma = \emptyset$. In this case, we have
     that there exists some $j$ such that $\varphi_j$ is satisfied by $\sigma$
     and $\Phi^*(c, \sigma) = \Phi^*(p_j, \sigma)$.
     Thus, $c$ is equivalent to $p_j$ with respect to $(\sigma, \delta)$.
     Since $count(p_j) < count(c)$, by the inductive hypothesis and
     transitivity of the equivalence relation, we conclude that
     there exists a plan $c'$ such that
     $c'$ is regressable with respect to $(\sigma,\delta)$ and $c'$
     is equivalent to $c$ with respect to $(\sigma,\delta)$.

     \item $Sens_a \setminus \sigma \ne \emptyset$.
Without loss of generality, we can assume that for each $\varphi_i
\in S$, there exists (at least) one $\sigma' \in \Phi(a,\sigma)$
such that $\varphi_i$ is satisfied in $\sigma'$.
     Using Lemma \ref{lemAdd3-son}, we can construct a plan
$c_1 = a; case\ (\varphi_1' \rightarrow p_1' , \ldots ,  \varphi'_n
\rightarrow
         p'_n)$
     which is potentially regressable and
     $\Phi^*(c_1,\sigma) =  \Phi^*(c, \sigma)$.
     From the construction of $c_1$, we know that for each
     $\sigma' \in \Phi(a,\sigma)$ there exists one and only one
     $j$, $1 \le j \le n$, such that $\varphi_j'$ is satisfied
     in $\sigma'$. Applying the inductive hypothesis
     for $(\sigma',\delta)$ and the plan $p_i'$,
     we know that there exists a regressable plan $q_i$
     which is equivalent to $p_i'$ with respect to $(\sigma',\delta)$.
     This implies that
     $c' = a; case\ (\varphi_1' \rightarrow q_1
         , \ldots ,  \varphi'_n \rightarrow q_n)$
     is equivalent to $c$ with respect to $(\sigma,\delta)$.
     Furthermore, every case plan in $c'$ is potentially regressable
     and each $q_i$ is regressable with respect to $(\sigma',\delta)$.
     To complete the proof, we will show that $c'$ is not
     redundant with respect to $(\sigma,\delta)$.
     Because
     for each $\sigma' \in \Phi(a,\sigma)$ there exists
     at most one $j$ such that $\varphi_j'$ is satisfied in
     $\sigma'$, none of the branches can be removed.
     Since $Sens_a \setminus \sigma \ne \emptyset$
    there are more than one a-state in
    $\Phi(a,\sigma)$.
     Therefore, we cannot replace $c'$ by one of its branches.
     This,
     together with the fact that $q_j$ is not redundant with respect to
     $(\sigma',\delta)$, implies
     that $c'$ is not redundant with respect to $(\sigma,\delta)$.
     The inductive hypothesis is proved for this case as well.
     \end{enumerate}

     \item $c = \alpha;c_1$ where $\alpha$ is a sequence of
     non-sensing actions and $c_1$ is a case plan.
     Let $P_\alpha = \{\alpha' \mid \alpha'$ is a subplan
     of $\alpha$ and there exists some $c_1''$ such that
     $\alpha';c_1''$ is equivalent to $c$ with respect to $(\sigma,\delta)\}$.
     Let $\beta$ be a member of $P_\alpha$
     such that $|\beta| = \min \{|\alpha'| \mid \alpha' \in P_\alpha\}$\footnote{
        For a sequence of actions $\gamma$, $|\gamma|$
    denotes the length of $\gamma$.
     }.
     Since $P_\alpha \ne \emptyset$, $\beta$ exists.
     We have that $\beta$ is a sequence of non-sensing actions,
     and so, $\Phi^*(\beta, \sigma) = \{\sigma_1\}$. It follows from the above
     case and the inductive hypothesis that
     there exists a regressable plan $c_1'$ which
     is equivalent to $c_1$ with respect to $(\sigma_1,\delta)$.
     Consider the plan $c' = \beta; c_1'$.
     We have that $c'$ is a  potentially regressable
     conditional plan. To complete the proof, we will show that
     $c'$ is not redundant with respect to $(\sigma,\delta)$.
     Assume the contrary, we will have three cases:
\begin{enumerate}[(a)]
\item There exists a subplan $\beta'$ of $\beta$
     such that $q = \beta';c_1'$ is equivalent to $c'$
     with respect to $(\sigma,\delta)$. This implies that
     $\beta';c_1$  is equivalent to $c'$
     with respect to $(\sigma,\delta)$ which contradicts the
     construction of $\beta$.
\item There exists a subplan $c''$ of $c_1'$
     such that $q = \beta;c''$ is equivalent to $c'$
     with respect to $(\sigma,\delta)$. This implies that
     $c''$  is equivalent to $c_1'$
     with respect to $(\sigma_1,\delta)$ which contradicts the
     construction of $c_1'$.
\item  There exists a subplan $\beta'$ of $\beta$ and
      a subplan $c''$ of $c_1'$
     such that $q = \beta';c''$ is equivalent to $c'$
     with respect to $(\sigma,\delta)$. This implies that
     $\beta' \in P_\alpha$ and $|\beta'| < |\beta|$, which is
    a contradiction on the construction of $\beta$.
     Thus this
     case cannot happen as well.
     \end{enumerate}
     This shows that $c'$ is not redundant with respect to $(\sigma,\delta)$.
     So, we have proved that $c'$ is regressable
     and equivalent to $c$
     with respect to $(\sigma,\delta)$.
     The inductive step is proved for this case.\qed
     \end{enumerate}

     \end{enumerate}

\begin{figure}[htb]
\begin{center}
\setlength{\unitlength}{1cm}
\begin{picture}(6,3.5)
\multiput(5.6,0)(-0.5,0){9}{\line(-1,0){0.3}}
\put(1,0){\vector(-1,0){0}}
\put(6,0){\makebox(0,0)[cc]{$_{\delta_G}$}}
\put(3,0.1){\makebox(0,0)[cb]{$_{{\mathcal R}^*(c',\delta_G)}$}}

\put(3,2.9){\makebox(0,0)[ct]{$_{\Phi^*(c,\sigma_I)}$}}

\put(0,0){\makebox(0,0)[cc]{$_{\delta \ne \bot, \sigma_I \in ext(\delta)}$}}
\put(0,3){\makebox(0,0)[cc]{$_{\sigma_I}$}}

\put(0.3,3){\vector(1,0){4.3}}

\multiput(3,2.3)(0,-0.5){3}{\line(0,-1){0.3}}
\put(3,0.8){\vector(0,-1){0}}
\put(3,2.3){\vector(0,1){0}}

\put(6,3){\makebox(0,0)[cc]{$_{
\Phi^*(c,\sigma_I) \subseteq ext(\delta_G)}$}}

\put(3,2.6){\makebox(0,0)[ct]{{\footnotesize c}}}
\put(3,3.1){\makebox(0,0)[cb]{\footnotesize Progression}}
\put(3.1,1.7){\makebox(0,0)[lc]{\footnotesize Equivalent w.r.t. $(\sigma_I,\delta_G)$}}
\put(3,-0.1){\makebox(0,0)[ct]{\footnotesize Regression}}
\put(3,0.5){\makebox(0,0)[cb]{{\footnotesize c'}}}

\end{picture}
\end{center}
     \caption{Illustration of Theorem \ref{theoAdd1-maintext}.}
     \label{example-fig-05}
\end{figure}
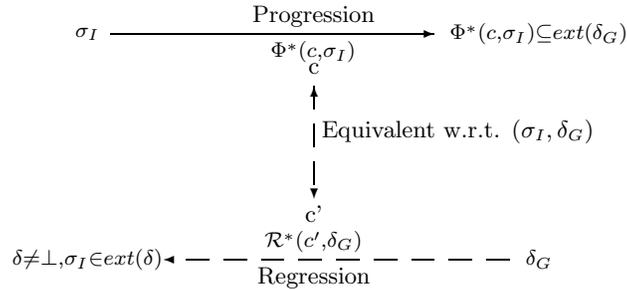

We are now ready to prove the completeness of our regression formulation,
which is illustrated by Figure \ref{example-fig-05}.

\begin{theorem} [Completeness of Regression] \label{theoAdd1-maintext}
Given a planning problem $P= \langle A,O,I,G \rangle$ and a
progression solution $c$ of $P$, there exists a regression solution
$c'$ of $P$ such that $c'$ is not redundant and is equivalent to $c$
with respect to $(\sigma_I, \delta_G)$.
\end{theorem}

\proof Lemma
\ref{lemAdd3} implies that there exists a regressable plan $c'$ with
respect to $(\sigma_I, \delta_G)$ which is equivalent to $c$ with
respect to $(\sigma_I, \delta_G)$. The non-redundancy of $c'$
follows from the fact that it is a regressable plan. The conclusion
of the theorem follows directly from Lemma \ref{lem31b} and Theorem
\ref{lem3-maintext}.\qed

\section{Related Work}
\label{related}

Waldinger \cite{Waldinger77} is probably the first to discuss
regression in Artificial Intelligence. In his paper, Waldinger uses
the concept of regression in \emph{plan modification}. To plan for
several goals simultaneously, say $P$ and $Q$, his strategy was to
first find a plan to achieve $P$, then modify that plan to achieve
$Q$. In order to achieve $Q$, regression is used to make sure that
any action added to the existing plan will not interfere with $P$.
Waldinger's regression is based on the idea of ``weakest
precondition'' proposed by Dijkstra in 1975 \cite{dijkstra75}(see
also, e.g., \cite{Best96,debakker90}). Intuitively, regression from
a logical sentence that is represented by a conjunction of goals,
$conj$, via an action, $A$, yields another logical sentence that
encodes what must be true before $A$ is performed to make $conj$
true immediately afterwards. This is computed by the formula
\[
S' = Prec(A) \cup (S \setminus Add(A)),
\]
where $S$ denotes the set of
goals in the conjunction $conj$, $S'$ denotes subgoals in the
regressed conjunction, $Pre(A)$ denotes the set of preconditions of
$A$, and $Add(A)$ denotes the set of add conditions of $A$;
something similar to what is proposed in \cite{weld94}. Following
Waldinger, Nilsson \cite{nilson80} discusses regression with respect
to partially grounded actions and proposes a regression algorithm
for plan generation.

Another early effort in formulating regression over simple
(non-sensing) actions is due to Pednault \cite{ped86}. In his Ph.D.
thesis \cite{ped86}, Pednault proposed the language $ADL$ (Action
Description Language) that extends STRIPS and allows, amongst other
things, conditional effects. In addition, Pednault also presents
sound and complete formula-based regression operators for $ADL$
actions. Addressing a similar problem, Reiter \cite{rei01b} also
presents a sound and complete formula-based regression formulation
over simple actions within the Situation Calculus framework. It
reduces reasoning about future situations to reasoning about the
initial situation using first-order theorem proving. Regression
operators are provided for the formulae, with and without functional
fluents.

Scherl and Levesque \cite{ScherlL03} were probably the first to
extend the regression formulation for simple actions to include
sensing actions. They directly formalize regression in first order
logic, within the framework of Situation Calculus. Their
formula-based regression operator is defined with respect to a set
of successor state axioms which was based on Moore's formulation of
accessible worlds \cite{moo85b}. They show that, for any plan $P$
expressed by a ground situation term $s_{gr}$ (a ground situation
term is built on the initial situation by repeatedly applying the
function $do$ on it), the axiomatization $F$ of a domain including
the successor state axioms $F_{ss}$, G is an arbitrary sentence then
\[F \models G(s_{gr})\ \iff\ F \setminus F_{ss} \models
R^*[G(s_{gr})],\] where $R^*(\varphi)$ indicates that the regression
operator is repeatedly applied until the regressed formulae is
unchanged. Intuitively, this shows that the regression is sound and
complete. However, Scherl and Levesque do not define regression over
conditional plans. Later, Reiter adapts the work of Scherl and
Levesque in his book \cite{rei01b}. He does not, however, consider
regression on conditional plans. De Giacomo and Levesque
\cite{deg99} consider a generalized action theory where successor
state axioms and sensing information are conditionally applicable.
For example the following conditional successor state axiom
\cite{deg99} expresses that if a robot is alone in a building, then
the status of the door is only defined by the robot's actions $open$
and $close$.

$Alone(s) \supset$\\
    \hspace*{0.6in} $DoorOpen(x,do(a,s) \equiv $\\
    \hspace*{0.9in} $a = open(x) \vee (a \neq close(x) \wedge DoorOpen(x,s)).$

Here the \emph{sensor fluent formula} $Alone(s)$ expresses the
condition that the robot is alone in the building in a situation
$s$, $DoorOpen(x,do(a,s)$ expresses the fact that a door $x$ is
open in the situation after the robot performs an action in the
situation $s$, and $DoorOpen(x,s)$ expresses the fact the door $x$
is open in the situation $s$. Similarly, the following conditional
sensed fluent axiom \cite{deg99} expresses the condition that if
the robot is outdoors, then its on-board thermometer always measures
the temperature around the robot.

$Outdoor(s) \supset$\\
    \hspace*{0.6in} $OutDoorTemperature(n,s) \equiv thermometer(s) = n$.

Their formula-based regression is then defined over
\emph{histories}. A history is defined as a sequence
$(\overrightarrow{v_0}).(A_1,\overrightarrow{v_1}), \ldots,
(A_n,\overrightarrow{v_n})$ where each $A_i$ is an action,
$\overrightarrow{v_i}$ represents a vector of the values $\langle
v_{i,1}, \ldots, v_{i,m}\rangle$ and $v_{i,j}$ represents the
reading value of $j^{th}$ sensor after the $i^{th}$ action.
However, they showed that the regression although sound, does
not guarantee completeness in some circumstances. They also did
not consider regression on conditional plans.

In another direction, Son and Baral \cite{sonbaral00} study
regression over sensing actions using the high-level action language
$\cala_K$. In this work, they provide a state-based transition
function and a formula-based regression function with respect to the
full semantics. Different from the work of \cite{deg99,rei01b}, Son
and Baral define regression over conditional plans. They also prove
that their regression formulation is both sound and complete with
respect to the transition function. However in \cite{sonbaral00},
Son and Baral do not consider precondition of actions in their
regression formulation.

The regression formalism presented in this paper differs from
earlier notion of regression for action theories with sensing
actions in \cite{rei01b,ScherlL03,sonbaral00} in that our definition
is a state-based regression formalism while the earlier definitions
are formula-based. With regards to regression on conditional plans,
we are not aware of any other work except  \cite{sonbaral00}. For
regression on non-sensing actions, our definition is close to the
formula used in \cite{bonet:geffner:hsp-01}.

\section{Conclusion, Discussion, and Future Work}
\label{conc}

In this paper, we developed a state-based regression function
in domains with sensing actions, incomplete information, and actions without
conditional effects.
We also extended the regression function to allow for the
regression over conditional plans.
We proved the soundness of the extended regression function
with respect to the definition of the progression function and
developed a relaxed notion of completeness for the regression
function.

It is interesting to note that for planning problems described in
this paper, the progression function developed in this paper is
equivalent to the full semantics for domains with sensing actions
and incomplete information and to the 0-approximation developed in
\cite{sonbaral00}. This implies that the regression function
$\mathcal R$ (and hence ${\mathcal R}^*$) is also complete with
respect to the full semantics for planning problems as defined in
Section \ref{pre}. Since the complexity of (conditional) planning
with respect to the 0-approximation is lower than that with respect
to the full semantics of sensing actions, this means that the
conditional planning problem for domains presented in this paper has
a lower complexity than it is in general. In other words, the
complexity of the conditional planning problem presented in this
paper in {\bf NP}-complete, whereas the complexity of the
conditional planning problem for action theories with conditional
effects is $\Sigma^2_P$-complete \cite{baral:ijcai99:aij}. We
observe that this complexity results are somewhat different than the
complexity results in \cite{Bylander94}, as the planning problems in
\cite{Bylander94} do not contain sensing actions and are complete.

It should be noted that the notion of a conditional plan in this paper
is not as general as in \cite{sonbaral00}. For example, we
do not consider plans of the form $c_1;c_2$ where $c_1$ and $c_2$ are
case plans. This is done to make the presentation of
the proofs easier to follow. Indeed, in \cite{Tuan04}, we proved that
all of the theorems in this paper are valid with respect to
conditional plans defined in \cite{sonbaral00}.

Finally, we would like to mention that we have developed
a regression-based planner, called {\bf CPR}, using the regression formulation
proposed in this paper \cite{TuanBZS05a}. The planner
employs the best first search strategy with a heuristic function
similar to the HSP-r heuristic function \cite{bonet:geffner:hsp-01}.
Due to the fact that most of the available benchmarks in planning
with sensing actions allow disjunction in the initial state and
conditional effects, an experimental evaluation of {\bf CPR}
against other planners could not be done with respect to the benchmarks.
We have therefore developed our own domains to test {\bf CPR}.
Our initial experimental result shows that {\bf CPR} performs reasonably
well \cite{TuanBZS05a}. The code of {\bf CPR} and the domains are
available at \url{http://www.cs.nmsu.edu/~tson/CPR}.

Our main goal in the near future is to extend the regression
formalism proposed in this paper to allow conditional effects
and disjunctive initial states. This will allow us to extend {\bf CPR}
to deal with conditional effects and to evaluate the planning approach
based on regression against forward chaining approaches.

\section*{Acknowledgment}
We are grateful to the anonymous referees whose useful comments
helped us improve this paper. We would also like to thank Tu Phan for his
comments on an ealier version of this paper which help us establish
Proposition 1.
A preliminary version of this paper appeared in \cite{regression-aaai04}.
Le-Chi Tuan and Chitta Baral were partially supported by NSF grants
0070463 and 0412000 and a ARDA/DTO contract.
Tran Cao Son was also partially supported by NSF grants CNS-0454066
EIA-0220590 and HRD-0420407.

\bibliographystyle{plain}
\bibliography{../../../bibtex/bibfile}

\end{document}